%% file: main_bio.tex
\documentclass{article}

\usepackage{arxiv}
\usepackage{natbib}
\input{packages}

\title{DeepCEL0 for 2D Single Molecule Localization in Fluorescence Microscopy}

\author{
	Pasquale Cascarano \\
	Department of Mathematics\\
	University of Bologna, 40126, Bologna, Italy\\
	\texttt{pasquale.cascarano2@unibo.it}
	\And
	Maria Colomba Comes\\
	Department of Electronic Engineering\\
	University of Tor Vergata, 00133, Rome, Italy \\
	\texttt{maria.colomba.comes@uniroma2.it}
	\And
	Andrea Sebastiani\\
	Department of Mathematics\\
	University of Bologna, 40126, Bologna, Italy\\
	\texttt{andrea.sebastiani3@unibo.it}
	\And 
	Arianna Mencattini \\
	Department of Electronic Engineering\\
	University of Tor Vergata, 00133, Rome, Italy \\
	\texttt{mencattini@ing.uniroma2.it}
	\And 
	Elena Loli Piccolomini \\
	Department of Computer Science and Engineering\\
	University of Bologna, 40126, Bologna, Italy\\
	\texttt{elena.loli@unibo.it} 
	\And
	Eugenio Martinelli \\
	Department of Electronic Engineering\\
	University of Tor Vergata, 00133, Rome, Italy \\
	\texttt{martinelli@ing.uniroma2.it}
	\And 
	
}

\begin{document}
\maketitle


%
%
%

\begin{abstract}
	In fluorescence microscopy, Single Molecule Localization Microscopy (SMLM) techniques aim at localizing with high precision high density fluorescent molecules by stochastically activating and imaging small subsets of blinking emitters. Super Resolution (SR) plays an important role in this field since it allows to go beyond the intrinsic light diffraction limit.\\
	In this work, we propose a deep learning-based algorithm for precise molecule localization of high density frames acquired by SMLM techniques whose $\ell_{2}$-based loss function is regularized by positivity and $\ell_{0}$-based constraints. The $\ell_{0}$ is relaxed through its Continuous Exact $\ell_{0}$ (CEL0) counterpart. The arising approach, named DeepCEL0, is parameter-free, more flexible, faster and provides more precise molecule localization maps if compared to the other state-of-the-art methods.  We validate our approach on both simulated and real fluorescence microscopy data. \\
	\textbf{Availability and implementation}: DeepCEL0 code is freely accessible at https://github.com/sedaboni/DeepCEL0
\end{abstract}

\section{Introduction}

The spatial resolution of images acquired by fluorescence microscopy refers to the shortest distance at which two fluorescent entities are perceived separately by the camera system. As a consequence of the light diffraction phenomena, lens with a uniformly illuminated circle aperture generate patterns known as Airy disks \citep{renz2013fluorescence}, namely the fluorescent emitters to be imaged are represented as blobs and not as isolated spots of light. The ability of the microscope to distinguish two relatively close entities is bounded by the well-known diffraction limit \citep{zheludev2008diffraction}, that represents an intrinsic constraint of the resolution of the optical acquisition device. More precisely, according to the Abbe's Criterion, the smallest resolvable distance by a light microscope corresponds roughly to half the optical wavelength, that is about 200 nm \citep{abbe1873beitrage}, thus compromising the direct observation of structures at nanoscale such as proteins, microtubules, mitochondria and less complex molecules. 

In the last decade, super-resolution microscopy techniques have revolutionized light microscopy biological imaging allowing biologists to see beyond the diffraction limit \citep{sahl2013super}. Among these techniques, in 2014, the pioneering works on STimulated Emission Depletion (STED) microscopy by Stefan W. Hell, and the development of Single-Molecule Localization Microscopy (SMLM) by Eric Betzig and William E. Moerner, have been awarded the Nobel Prize in Chemistry.   

In particular, SMLM techniques, such as Photo-Activated Localisation Microscopy (PALM) \citep{betzig2006imaging,hess2006ultra} and STochastic Optical Reconstruction Microscopy (STORM) \citep{rust2006sub}, provide high-precision molecule localization by sequentially activating and imaging a small percentage of photoswitchable fluorophores (emitters). \\
These molecules absorb and emit light of a specific wavelength. Therefore, by imaging a sparse set of activated emitters at a time, it is possible to identify their precise location in the Field of View (FOV). First, these SMLM techniques provide a stack of diffraction-limited frames, containing blobs (Airy disks), modelled by Gaussian Point Spread Functions (PSFs) \citep{rossmann1969point}. Then, each frame is analyzed separately with the aim of providing high precision localization maps of the emitters. After the individual processing is performed, all the frames are re-combined together to finally obtain a unique super-resolved image overcoming the diffraction limit. 

The sparser the set of activated emitters per frame is, the more precise the localization is. However, considering sparse frames takes a longer acquisition time thus limiting the ability to capture fast dynamics within live specimens.
Conversely, a high density of activated emitters negatively affects the quality of the super-resolved image in terms of localization precision. Indeed, localization in high-density settings, that are characterized by overlapping PSFs, represents a challenging task for all the existing sophisticated localization software tools \citep{sage2015quantitative,bernhem2018smlocalizer,davis2020rainbowstorm}. 

Approaches based on diverse rationale, from blinking statistics \citep{cox2012bayesian,dertinger2009fast,gustafsson2016fast} and standard Gaussian fitting or centroid estimation \citep{henriques2010quickpalm}, to subtraction of the model PSF \citep{gordon2004single,serge2008dynamic,qu2004nanometer} or deconvolution with sparsity-promoting constraints \citep{holden2011daostorm,hugelier2016sparse,gazagnes2017high,solomon2019sparcom,min2014falcon}, have been proposed to handle high-density data.

According to the deconvolution approaches with sparsity-promoting priors, for each frame, the emitters localization maps are viewed as the critical points of a least squares $\ell_0$-penalized objective. In particular, in \citep{soubies2015continuous}, the authors define the Continuous Exact $\ell_0$ (CEL0) regularizer, which has been proven to be effective in the field of SMLM with high-density data \citep{gazagnes2017high}. This CEL0-based method solves, through an iterative scheme, a continuous non-convex optimization problem whose objective function is a weighted sum of an $\ell_{2}$ fidelity term and the CEL0 regularizer. Furthermore, positivity constraints are also added to the model since molecules locations must be reconstructed. Despite its state-of-the-art performances in retrieving the localization maps, the accuracy of the method strictly depends on the choice of the parameter balancing the $\ell_{2}$ and the sparsity-promoting terms, thus drastically limiting its usage in real experiments. 

More recently, a deep learning-based approach that exploits a Convolutional Neural Network (CNN) has been developed \citep{nehme2018deep}. The method, called DeepSTORM, is trained on synthetic (artificially generated) frames, and tested directly on experimental data, thus avoiding to collect a huge amount of training samples related to the particular experiment under study. Despite being a fast and parameter-free super-resolution microscopy algorithm able to manage high-density data, DeepSTORM is not designed to reconstruct high precision localization maps.

In the present paper, we propose a method tailored to perform the localization, named DeepCEL0, that bridges the gap between deep learning-based and regularized deconvolution approaches. Basically, we exploit the skeleton of the network architecture underneath DeepSTORM using the CEL0 penalty as part of the loss function for training the net. We further add positivity constraints through the insertion of a RELU layer in the network architecture. The strength of DeepCEL0 consists in embedding and joining the main advantages of the two standard methods, thus supplying, on the one hand, a quite fast and parameter-free deep-learning algorithm, like DeepSTORM, and, on the other hand, an efficient algorithm providing high precision localization maps, like CEL0. We report quantitative and qualitative molecule localization results on the simulated and real datasets from IEEE ISBI SMLM challenge \footnote{\url{http://bigwww.epfl.ch/smlm/}}. We also compare the performances achieved by DeepCEL0 with those obtained by the two baseline state-of-the-art methods, CEL0 and DeepSTORM. 

\input{method}
\input{results}

%
%

\section{Conclusion}

In this paper, we present a deep learning-based method called DeepCEL0 for precise single molecule localization in high density fluorescence microscopy settings. The proposed method brings together the benefits of two well-known standard methods in the field, i.e., DeepSTORM and CEL0, introducing a network architecture with two main novelties: a continuous $\ell_{0}$-penalized training loss function and the adoption of positivity constraints on the solution through a ReLu layer. Compared to the standard methods, numerical results show how DeepCEL0 can provide very high precision localization maps, without detriment to computational cost. Moreover, the method is parameter-free and can be easily tested and applied on real data after a training phase on only synthetic data. The promising results make the methods easy to perform in disparate real applications exploiting fluorescence microscopy.



\section*{Funding}
P.C. and E.L.P. have been partially supported by the GNCS-INDAM project 2020 \textit{"Ottimizzazione per l’apprendimento automatico e apprendimento automatico per l’ottimizzazione".}

\input{main_bio.bbl}

\end{document}

%% file: packages.tex
\usepackage{amsmath,amsfonts,amssymb}
\usepackage{bm} 
\usepackage{cases}
\usepackage{float}
\usepackage{mwe}
\usepackage{mathtools}
\usepackage{comment}
\usepackage{empheq}
\usepackage{booktabs}
\usepackage{textcomp}
\usepackage{algorithm,algorithmic}
\usepackage{array}
\usepackage{subfig}
\usepackage{url}
\usepackage{caption}
\usepackage{multirow}
\usepackage{systeme}
\usepackage{soul}
\usepackage{framed,multirow}
\usepackage{amssymb}
\usepackage{latexsym}
\usepackage{lineno}
\usepackage{multicol}
\usepackage{booktabs}
\usepackage{array}
\makeatletter
\newcommand{\thickhline}{%
    \noalign {\ifnum 0=`}\fi \hrule height 1pt
    \futurelet \reserved@a \@xhline
}
\newcolumntype{"}{@{\hskip\tabcolsep\vrule width 1pt\hskip\tabcolsep}}
\makeatother

\usepackage{tikz}
\usetikzlibrary{positioning}
\usetikzlibrary{3d}
\usetikzlibrary{matrix}
\usetikzlibrary{decorations.text}
\usetikzlibrary{spy}

\makeatletter
\tikzoption{canvas is plane}[]{\@setOxy#1}
\def\@setOxy O(#1,#2,#3)x(#4,#5,#6)y(#7,#8,#9)%
{\def\tikz@plane@origin{\pgfpointxyz{#1}{#2}{#3}}%
	\def\tikz@plane@x{\pgfpointxyz{#4+#1}{#5+#2}{#6+#3}}%
	\def\tikz@plane@y{\pgfpointxyz{#7+#1}{#8+#2}{#9+#3}}%
	\tikz@canvas@is@plane
}
\makeatother
\tikzoption{canvas is xy plane at z}[]{%
	\def\tikz@plane@origin{\pgfpointxyz{0}{0}{#1}}%
	\def\tikz@plane@x{\pgfpointxyz{1}{0}{#1}}%
	\def\tikz@plane@y{\pgfpointxyz{0}{1}{#1}}%
	\tikz@canvas@is@plane
}
\makeatother

\usepackage[english]{babel}
\usepackage{tipa}
 \expandafter\let\csname equation*\endcsname\relax
 \expandafter\let\csname endequation*\endcsname\relax
 
\definecolor{green_mat}{rgb}{0.01,0.75,0.24}
\definecolor{purple_mat}{rgb}{0.49,0.18,0.56}
\definecolor{blue_mat}{rgb}{0.00,0.45,0.74}

\newcommand{\Kbold}{\mathbf{K}}

\newcommand{\Hbold}{\mathbf{H}}

\newcommand{\Sbold}{\mathbf{S}}

\newcommand{\Xbold}{\mathbf{X}}
\newcommand{\Ybold}{\mathbf{Y}}

\newcommand{\Zbold}{\mathbf{Z}}

 \newcommand{\cbold}{\mathbf{c}}

 \newcommand{\xbold}{\mathbf{x}}

%% file: method.tex
\section{Mathematical Background}

\subsection{A mathematical image formation model for PALM/STORM acquisitions}
We now recast the problem of finding high precision localization maps as a standard image super resolution inverse problem. The image acquisition model we develop is used in the experimental section to construct reliable synthetic data simulating PALM/STORM diffraction-limited acquisitions. Let $\Ybold \in \mathbb{R}_{+}^{M \times M }$ be a frame acquired by PALM/STORM techniques representing a sparse set of activated molecules on a coarse pixel-grid of dimension $M \times M$. Let $\Xbold \in \mathbb{R}_{+}^{N \times N}$ be the localization map referred to the diffraction-limited PALM/STORM acquisition, that is a HR image defined on a $L$-time thinner pixel-grid such that $N=LM$, with $L$ a non-negative integer termed as magnification factor.

In the following, for the sake of the readability of the manuscript, given a generic 2D image $\Zbold$ of dimension $m \times n$ we denote by $\overrightarrow{\Zbold}$ its vectorized version of dimension $mn \times 1$.

The acquisition of the LR image $\overrightarrow{\Ybold}$ is modelled through the following discrete degradation process:  

\begin{equation} \label{eq:inverse_problem2}
    \overrightarrow{\Ybold} = \mathcal{N}(\Sbold_{L} \Kbold \overrightarrow{\Xbold}),
\end{equation}

where $\mathcal{N}$ models the presence of Poisson noise and Additive White Gaussian (AWG) noise, which are typical aberrations in fluorescence imaging \citep{garini1999signal,waters2009accuracy,jezierska2012poisson}. \\
In order to simplify the model, we remark that the Poisson noise is a data dependent noise which can be well approximated as AWG noise. Therefore, in the following, we restrict to AWG as noise aberration affecting the LR image $\overrightarrow{\Ybold}$, and then: 

\begin{equation} \label{eq:inverse_problem}
    \overrightarrow{\Ybold} = \Sbold_{L} \Kbold \overrightarrow{\Xbold} + \eta,
\end{equation}
where $\eta \in \mathbb{R}^{M^{2}}$ is a realization of a Gaussian random variable with mean zero and standard deviation $\sigma_{\eta}$.

The presence of the typical Airy disk patterns (blobs) in the LR acquisition is modelled through Gaussian PSFs. Therefore, in our model (Eq. \eqref{eq:inverse_problem}) we assume the operator $\Kbold \in \mathbb{R}^{N^{2} \times N^{2}}$ denotes the discretization of a convolution with a Gaussian kernel $\textit{k} : \mathbb{R}^{2} \to \mathbb{R}$ with mean equal to zero and standard deviation $\sigma_{\textit{k}}$, which is defined as follows: 

\begin{equation} \label{eq:PSF}
    \textit{k} (x,y) := \dfrac{1}{\sigma_{\textit{k}} \sqrt{2 \pi}} \text{exp} \left( - \dfrac{x^{2}+ y^{2}}{2\sigma_{\textit{k}}^{2}}\right).
\end{equation}
 
The operator $\Sbold_{L} \in \mathbb{R}^{M^{2} \times N^{2}}$ in Eq. \eqref{eq:inverse_problem} refers to the discretization of the downsampling operator linking the HR localization map represented on a fine pixel grid to the PALM/STORM acquired image defined on a $L$-time coarser pixel grid.

\subsection{A deconvolution approach with a sparsity-promoting prior for single molecule localization}

The problem in Eq. \eqref{eq:inverse_problem} is well-known to be ill-posed, meaning that, given $\overrightarrow{\Ybold}$, it is not feasible to retrieve $\overrightarrow{\Xbold}$ by simply inverting the degradation described by $\Sbold_{L}\Kbold$, due to the lack of uniqueness and stability of the solution. 
Since, for each frame, the related localization map is sparse, a quite common approach to overcome this instability, consists in forcing the solution to satisfy some sparsity constraints through the $\ell_{0}$ functional. More precisely, an estimate $\overrightarrow{\Xbold}^{*}$ of the unknown $\overrightarrow{\Xbold}$ can be viewed as the solution of the following $\ell_{0}$-regularized optimization problem: 

\begin{equation} \label{eq:l0-optimization}
    \overrightarrow{\Xbold}^{*} \in \underset{\xbold \in \mathbb{R}^{N^2}}{\arg \min} \dfrac{1}{2} \lVert \Sbold_{L} \Kbold \xbold - \overrightarrow{\Ybold} \rVert^{2}_{2} + \lambda \lVert \xbold \rVert_{0} + \mathbf{1}_{\geq 0} (\xbold),
\end{equation}

where $\lVert \cdot \rVert_{0} : \mathbb{R}^{N^2} \to \mathbb{R}$ denotes the $\ell_{0}$ functional defined as: 

\begin{equation} \label{eq:l0_penalizer}
\lVert \xbold \rVert_{0}:=\sum_{i=1}^{N^2} | \xbold_{i} |_{0} \quad    \text{with} \quad                | \xbold_{i} |_{0}:= \begin{cases}
                        0 & \xbold_{i} = 0  \\
                        1 & \xbold_{i} \neq 0,
                          \end{cases}
\end{equation}

which counts the number of non-zero elements of $\xbold \in \mathbb{R}^{N^2}$. Moreover, $\lambda >0$ is the trade-off parameter measuring the level of sparsity of $ \overrightarrow{\Xbold}^{*}$, whereas $\mathbf{1}_{\geq 0}(\cdot)$ is formally the characteristic function of the positive octant of $\mathbb{R}^{N^2}$ constraining the computed estimation $\overrightarrow{\Xbold}^{*}$ to have positive entries and $\lVert \cdot \rVert_{2}$ is the $\ell_{2}$-norm.\\ 
In \citep{gazagnes2017high} the authors propose to replace the $\ell_{0}$ penalization \eqref{eq:l0_penalizer} by its continuous relaxation $\Phi_{\text{CEL0}} : \mathbb{R}^{N^2} \to \mathbb{R}$, named as CEL0 penalizer \citep{soubies2015continuous}, which reads:  

\begin{align}
    \Phi_{\text{CEL0}}(\xbold) := \sum_{i=1}^{N^{2}} \lambda_{\text{CEL0}} -  \dfrac{\lVert \cbold_{i} \rVert}{2} \left( |\xbold_{i}| - \dfrac{\sqrt{2\lambda_{\text{CEL0}}}}{\lVert \cbold_{i} \rVert} \right)^{2} \mathbf{1}_{V_{i}}, 
\end{align}

where by $\mathbf{1}_{V_{i}}$ we denote the characteristic function of the set $V_{i}:= \lbrace \xbold_{i}\in\mathbb{R} \big{|} \lvert\xbold_{i}\rvert < \frac{\sqrt{2\lambda_{\text{CEL0}}}}{\lVert \cbold_{i} \rVert} \rbrace$; by $c_{i}$ we denote the $i$-th column of the matrix $\Sbold_{L} \Kbold$ for $i=1 \dots N^{2}$, whereas the positive scalar $\lambda_{\text{CEL0}}$ balances the strength of the sparsity induced by the CEL0 penalizer. 
Therefore, given a PALM/STORM acquisition $\overrightarrow{\Ybold}$, the CEL0-based method estimates the high precision localization map $\overrightarrow{\Xbold}^{*}$ by solving the following optimization problem:

\begin{equation} \label{eq:CEL0_minimization}
\overrightarrow{\Xbold}^{*} \in \underset{\xbold \in \mathbb{R}^{N^2}}{\arg \min} \dfrac{1}{2} \lVert \Sbold_{L} \Kbold \xbold - \overrightarrow{\Ybold} \rVert^{2}_{2} + \Phi_{\text{CEL0}}(\xbold) + \mathbf{1}_{\geq 0} (\xbold).
\end{equation}

For its numerical solution, in \citep{gazagnes2017high} the authors make use of the iterative reweighted $\ell_{1}$ (IRL1) strategy \citep{ochs2015iteratively} which is tailored to handle nosmooth noconvex optimization problems.

\subsection{A deep learning based method for single molecule localization}

In \citep{nehme2018deep} the authors have introduced a fast and parameter-free deep learning-based method termed as DeepSTORM, which makes use of a CNN to provide a HR counterpart of the LR frames acquired by PALM/STORM techniques. In particular, given an outer training set of $K$ 2D-image pairs $\lbrace \left(\Ybold_{k}, \Xbold_{k}\right) \rbrace_{k=1 \dots K}$, where $\Ybold_{k} \in \mathbb{R}^{M \times M}$ denotes the LR input and $\Xbold_{k} \in \mathbb{R}^{N \times N}$ denotes its HR counterpart, and a particular encoder-decoder architecture $f_{\theta}$ with weights $\theta$, DeepSTORM training involves the following $\ell_{1}$-regularized loss function:

\begin{equation} \label{eq:loss_deepStorm}
    \dfrac{1}{K} \left( \sum_{k=1}^{K} \lVert g \ast f_{\theta}(\text{NN}(\Ybold_{k})) - g \ast \Xbold_{k} \rVert^{2}_{F} + \lVert \overrightarrow{f_{\theta}(\text{NN}(\Ybold_{k}))} \rVert_{1} \right),
\end{equation}

where by NN we denote the coarse Nearest Neighbour super resolution algorithm resampling a LR image on a $L$-times finer grid by simply replicating the pixel values; by $\ast$ we refer to the discrete convolution with a 2D Gaussian kernel $g$ whose standard deviation is equal to 1. Moreover, by $\lVert \cdot \rVert_{F}$ and  $\lVert \cdot \rVert_{1}$ we denote the Frobenius norm and the $\ell_{1}$-norm, respectively. 

Once provided the final set of trained weights $\theta^{*}$ and a PALM/STORM acquisition $\Ybold$, an approximation $\Xbold^{*}$ of the HR localization map $\Xbold$ is obtained by computing $f_{\theta^{*}}(\text{NN}(\Ybold))$. 

Roughly speaking, DeepSTORM approximates the map linking the coarse Nearest Neighbour HR approximations of the PALM/STORM LR frames to their high-quality HR counterparts. 

As far as the training set is concerned, one of the main novelties of DeepSTORM is that it can provide good performances on real images even if it is trained only on a synthetic dataset created using an ImageJ plug-in called ThunderStorm \citep{ovesny2014thunderstorm}.


\section{Method}

\subsection{A CEL0 regularized loss function}

Our proposal combines the learning-based approach DeepSTORM and the sparsity-constrained deconvolution approach CEL0. In particular, we aim at providing a method which preserves the main advantages of the both aforementioned approaches, namely the ability of CEL0 to retrieve high precision localization maps and the fast and parameter-free computation provided by DeepSTORM. Therefore, following the idea of \citep{nehme2018deep}, we could train a CNN architecture $f_{\theta}$ based on the following regularized loss function: 

\begin{equation} \label{eq:loss_storm_CEL0}
    \dfrac{1}{K} \Bigg{(}\sum_{k=1}^{K} \lVert g \ast f_{\theta}(\text{NN}(\Ybold_{k})) - g \ast \Xbold_{k} \rVert^{2}_{2} +  \Phi_{\text{CEL0}}(\overrightarrow{f_{\theta}(\text{NN}(\Ybold_{k}))}) + \mathbf{1}_{\geq 0} (\overrightarrow{f_{\theta}(\text{NN}(\Ybold_{k}))})\Bigg{)},
\end{equation}

where $\lbrace \left(\Ybold_{k}, \Xbold_{k}\right) \rbrace_{k=1 \dots K}$ is a prefixed training set of $K$ LR-HR 2D-image pairs of size $M\times M$ and $N \times N$, respectively, whilst NN referrers to the Nearest Neighbour upscaling of factor equal to $L$.   

Differently from \citep{nehme2018deep}, the loss in Eq. \eqref{eq:loss_storm_CEL0} presents two main differences:  the $\ell_{1}$-regularizer is replaced by the CEL0 regularizer in order to guarantee better sparse reconstructions and the characteristic function of the positive octant is supplied to enforce positivity constraints on the computed solution. Unfortunately, the latter leads to a noncontinuous functional which can be difficult to train. To overcome this issue, we neglect the characteristic function term, thus training $f_{\theta}$ w.r.t. the following loss function: 

\begin{equation} \label{eq:loss_storm_CEL02}
    \dfrac{1}{K} \Bigg{(}\sum_{k=1}^{K}  \lVert g \ast f_{\theta}(\text{NN}(\Ybold_{k})) - g \ast \Xbold_{k} \rVert^{2}_{2} +  \Phi_{\text{CEL0}}(\overrightarrow{f_{\theta}(\text{NN}(\Ybold_{k}))})\Bigg{)}. 
\end{equation}

To force positivity constraints on the reconstructed solutions, we consider a CNN architecture $f_{\theta}$ which is a slightly modified version of the CNN architecture used in \citep{nehme2018deep}. In the following paragraphs, we describe in detail the skeleton of the architecture and the synthetic training set considered in our experimentation.


\subsection{A positivity-promoting Deep Architecture}

The considered CNN architecture, denoted as $f_{\theta}$ in Eq. \eqref{eq:loss_storm_CEL02}, is a modified version of the one initially proposed in \citep{nehme2018deep}. For the sake of brevity, we call convolutional layer the composition of convolutional filters with a batch normalization layer \citep{ioffe2015batch} followed by a ReLU non-linearity \citep{glorot2011deep} as activation function. The original architecture is an encoder-decoder and it is composed of seven convolutional layers. Each layer uses $3\times 3$ kernels of different depth equal to 32, 64, 128 and 512, respectively. In the encoder part, the filters' depth increases, and a $2\times 2$ max-pooling is used as downscaling operator to compress the features. In the decoder part, the layers are interleaved with a nearest neighbour upsampling operator and the filters' depth decreases. At the end of the network, another layer is added to compute the pixel-wise prediction. This layer is a $1\times 1$ convolutional filter. In the original implementation, this last layer uses a linear activation function, i.e., the identity. In order to induce positivity constraints to the computed solution, we replace the activation layer with ReLU.

\subsection{Synthetic training set and implementation notes}

In the experimental section, we evaluate how our method performs when dealing with an upsampling factor $L$ equal to $4$. According to this choice, we now describe the considered synthetic training set $\lbrace \left(\Ybold_{k}, \Xbold_{k}\right) \rbrace_{k=1 \dots K}$. In the following, we refer to $\Ybold_{k}$ as the input image and to $\Xbold_{k}$ as the target image, respectively. 
As well as for DeepSTORM \citep{nehme2018deep}, we generate a synthetic dataset made up of $20$ high density images containing 6 emitters per $\mu m^{2}$. The emitters are positioned on a FOV of size $64 \times 64$ pixels such that each pixel has size of 100 nm. We extract from these high density images $K=10000$ patches of size $26\times 26$. By projecting the emitter positions on a 4-times thinner pixel grid, we build the target images $\Xbold_{k}$ of size $104 \times 104$. 

The input images are constructed accounting for the corruptions affecting the acquired diffraction-limited data to be reconstructed, that are the standard deviation of the Gaussian PSF modelling the Airy patterns $\sigma_{\textit{k}}$ and the noise standard deviation $\sigma_{\eta}$. Therefore, $\Ybold_{k}$ for $k= 1 \dots K$ are obtained by blurring the $26\times 26$ patches with a discrete Gaussian kernel and by adding noisy components. Finally, these corrupted synthetic patches are upsampled by a factor equals 4 through the NN interpolation algorithm, and then used as input of the network. \\ 
We stress that, in real applications, the standard deviation of the PSF considered, if unknown, can be estimated using the Abbe's criterion which requires the light wavelength and the numerical aperture of the optical device used for acquiring the experimental data under study. Furthermore, the amount of noise can be either calculated directly from the microscope and detector characteristics or even through several mathematical techniques (\textit{see} \citep{paul2010automatic,mandracchia2020fast} and references therein). 

We use this synthetic dataset to train the proposed DeepCEL0 by minimizing the loss function defined in Eq. \eqref{eq:loss_storm_CEL02}. We train the network for $100$ epochs on batches containing $16$ samples using Adam optimizer and setting the learning rate equal to $0.001$. \\
In \eqref{eq:loss_storm_CEL02}, we set $g$ as a fixed Gaussian kernel, whose standard deviation equal to 1. Moreover, we point out that the contribution of the CEL0 regularization term can be weighted using the parameter $\lambda_{\text{CEL0}}$. It is worth noting that in this deep learning-based framework the parameter $\lambda_{\text{CEL0}}$ has not the same meaning of the regularization parameter in the deconvolution approach. \eqref{eq:CEL0_minimization}. Indeed, in the former case it does not directly correspond to a degree of sparsity of the computed solution. In our experiments, we tested different values for $\lambda_{\text{CEL0}}$ but we observed that setting it to $100$ provides outstanding results for all the tests.  The network is implemented in TensorFlow and the code is available at https://github.com/sedaboni/DeepCEL0. 


%% file: results.tex
\section{Results}

\subsection{Performance Evaluation and competitors} \label{sec:Performance Evaluation}

We assess the quality of the high resolution localization maps provided by our method and the competitors both on synthetic and real PALM/STORM images through quality metrics and visual inspections. When the ground-truth (GT) images are available, the performances are evaluated by pairing the GT molecules with the estimated ones: a match between a GT and an estimated molecule is created when the distance between their localizations is lower than a set tolerance $\delta$, whose standard values, if expressed in terms of pixels, are 2, 4 and 6 \citep{sage2015quantitative}. Such a tolerance $\delta$ is chosen lower than the Full-width at Half Maximum (FWHM) of the estimated Gaussian PSF modelling the Airy pattern. In the following, the matched estimated molecules up to the given tolerance are defined as True Positive (TP) molecules; the remaining estimated molecules are referred to as False Positive (FP) molecules; and finally, the GT molecules with no match are categorized as False Negative (FN) molecules. Beyond the estimated molecules, it is important to take into account the pixels of the GT images not corresponding to any molecule, which are labelled as True Negative (TN) molecules.  
The performances are assessed by computing the following evaluation metrics:
\begin{align}
     \text{Jaccard} (\%) & =  \frac{\text{TP}}{\text{TP}+\text{FP}+\text{FN}} \times 100, \\
     \text{Sensitivity}   (\%) & = \frac{\text{TP}}{\text{TP} + \text{FN}} \times 100, \\
     \text{Specificity}   (\%) & = \frac{\text{TN}}{\text{TN} + \text{FP}} \times 100. 
\end{align}

In order to quantify the level of corruption in the simulated data, we consider the signal-to-noise ratio (SNR) given by the following formula:

\begin{equation}
    \text{SNR}\text{(dB)}:= 10 \log_{10} \left( \dfrac{\lVert \Sbold \Hbold \overrightarrow{\Xbold}\rVert_{2}^{2}}{\lVert \Sbold \Hbold \overrightarrow{\Xbold} - \overrightarrow{\Ybold} \rVert_{2}^{2}} \right)
\end{equation}

where $\overrightarrow{\Xbold}$ is the real vectorized molecule localization map and $\overrightarrow{\Ybold}$ is the LR counterpart. 

We compare our approach with the state-of-the-art algorithms CEL0 and DeepSTORM on both high density synthetic and real SMLM PALM/STORM low resolution images whose level of corruption ranges from SNR = 15dB to SNR = 10 dB.

\subsection{Localization on synthetic test images}

DeepSTORM algorithm could provide a fast SR image reconstruction, although it is not tailored to provide high precision localization maps of the emitters. Conversely, CEL0 is designed to localize the molecules but presents two main flaws: the method is strongly dependent on the choice of the regularization parameter and the overall computation process is largely slower than the one provided by DeepSTORM.\\
Once the network has been trained, the proposed DeepCEL0 and DeepSTORM share a comparable computational time for SR image reconstruction. Moreover, as we demonstrate in the following, DeepCEL0 is also able to effectively localize the molecules.

First of all, to validate the localization performances of our proposal, we construct three different synthetic test scenarios as GT images, simulating high resolution fluorescence microscopy localization maps on a FOV of $512 \times 512$ pixels of size 25 nm. The two first scenarios, referred as Test 1a and Test 2a, represent two molecules, arranged on the 256th column, at distance of 25 nm and 75 nm, respectively. The other scenario, referred as Test 3a, shows four molecules disposed on a circle of a radius equal to 125 nm. Two molecules are arranged on the 256th column whereas the others are arranged on the 256th row. Once set $L=4$, by down-sampling the GT images according to the model in Eq. \eqref{eq:inverse_problem}, we simulate the acquisition of $128 \times 128$ LR diffraction-limited frames, where the PSF is modelled by a Gaussian function whose FWHM is equal to 258.21 nm, that is $\sigma_{\textit{k}} \approx $ 110 nm. 

In the lower panel of Fig. \ref{fig:sintetic_reconstruction}, we report the close-ups (x10 zooming) of the Region of Interest (ROI), namely the central zone of the $512 \times 512$ images where the synthetic spots are located, for the GT and Nearest Neighbour (NN), DeepSTORM, DeepCEL0 and CEL0 super resolved reconstructions. We remark that all the reconstructions reported have been normalized in the interval [0,1].\\
The zooms related to these methods are highlighted by blue, purple, green, red and yellow boxes, respectively. It is noteworthy that in the purple box the NN $\times 4$ upsampled image is reported in order to visualize the LR image at the same dimensions of the GT image. For the CEL0 method, we depict two different reconstructions obtained by setting two different regularization parameters for each test case.  \\
In the upper panel of Fig. \ref{fig:sintetic_reconstruction}, we draw the line profiles corresponding to the 256th column of the $512\times512$ GT, NN and DeepCEL0 images. 

For all the three tests, the two molecules arranged on the 256th column correspond to the two peaks in the line profile of the GT image (\textit{see} the blue line and blue box), whereas we observe they are completely overlapped and indistinguishable in the diffraction limited NN upsampled image's line profile (\textit{see} purple line and purple box).

As qualitative results, in Test 1a, Test 2a and Test 3a, DeepCEL0 is able to well approximate the positions of the two molecules (\textit{see} the red dashed line and red box). In all the three tests considered, the synthetic molecules are placed at a distance which is largely smaller than FWHM. In particular, Test 3a is the most challenging scenario since the number of molecules to estimate is greater than two. Also in this case, DeepCEL0 provides broadly better performances than DeepSTORM and more stable performances with respect to the CEL0 method. 
Indeed, DeepSTORM (\textit{see} green box) does not separate the two synthetic molecules and provides a large number of FP molecules, thus confirming it is not designed to produce accurate localization maps \citep{nehme2018deep}. 
CEL0, as expected, provides highly accurate localization maps, but, as we can observe, its performances are highly sensitive to the choice of the regularization parameter. Furthermore, it is remarkable how small changes of the regularization parameter produce largely different reconstructions and how different scenarios require different optimal parameters. 

Finally,  the results provided by DeepCEL0 have been computed by using the same architecture trained setting $\lambda_{\text{CEL0}}=100$. Therefore, these tests prove the effectiveness of our approach retrieving highly precise localization maps and more stability with respect to the choice of the hyperparameter $\lambda_{\text{CEL0}}$, if compared to the CEL0 method.   

\subsection{Localization on IEEE ISBI simulated dataset}

We now consider a more realistic dataset provided by the 2013 IEEE ISBI SMLM challenge. The dataset is a stack of 361 different frames used as GT localization maps in our analysis. A total number of 81049 emitters are counted: for each frame 217 fluorophores are activated on average. The frames simulate realistic high density acquisitions of 8 tubes of diameter size equal to 30 nm depicted on a FOV of $256 \times 256$ pixels of size 25 nm. We simulate LR diffraction-limited acquisitions by applying the image formation model in Eq. \eqref{eq:inverse_problem} setting $L=4$, thus representing the GT scenarios on a coarser grid of $64 \times 64$ pixels of size 100 nm. The PSF is modelled by using a Gaussian function of FWHM = 258.21 nm ($\approx$ $\sigma_{\textit{k}} = 110$ nm). We further corrupt the LR frames by adding three different realization of Gaussian noise, such that SNR = 10 dB, 12 dB, 15 dB, respectively referred to as Test 2b, Test 3b, Test 4b. Finally, we consider a more realistic scenario, referred to as Test 1b, where the frames are corrupted by Gaussian noise components of different standard deviations. 

The localization performances of CEL0, DeepSTORM and DeepCEL0 are assessed for the four tests by using the evaluation metrics introduced in Section \ref{sec:Performance Evaluation}. In Table \ref{tab:1}, we report the results expressed in terms of Jaccard index for $\delta$ set as 2,4,6 pixels, corresponding to a tolerance of 50,100,150 nm, respectively. The sensitivity and specificity values, instead, are computed only with respect to the tolerance $\delta = 2$, which expresses a more faithful compliance of the reconstructions compared to the GT.  

DeepSTORM and DeepCEL0 models are trained on synthetic images (\textit{see} Sub-Section 3.3) such that the downsampling factor $L$ and the PSF are set as above and the Gaussian noise corruption leads to an SNR value equals 15 dB on average. These two trained models are used for all the four tests considered. For the sake of a fair comparison, we estimate the CEL0 regularization parameter by a trial and error procedure sampling 30 values over the range $[1e-3,1]$ on randomly chosen 8 frames among the stacked ones. In the following, we refer to the best regularization parameter for CEL0 as the one maximazing the Jaccard index with tolerance $\delta$ equals 2. 

In Table \ref{tab:1}, we report for each test (Test 1b, 2b, 3b, 4b) the results of CEL0, DeepSTORM and DeepCEL0. For what concerns CEL0, in the first row referred to each test, we draw the results obtained with respect to the best regularization parameters, whereas in the second row we show the results obtained by choosing as regularization parameter the best one obtained for Test 4b.  

We inspect in depth Test 2b, the most challenging one due to a low SNR value. In Fig. \ref{fig:localization}, we provide the localization maps of the reconstructions for the frame 2.  Green circles indicate the TP molecules, whereas red crosses highlight the FP molecules. In Fig. \ref{fig:ISBI}, we report the sum of all the stacked reconstructed SR frames provided by the competing methods and of the stacked GT and LR frames. In particular, for GT and all the methods, above the magenta line we report the activated molecules as white spots, whereas below the magenta line we depict the normalized images. The three close-ups show three region of interest where the tubulins seems completely overlapped in the LR acquisition because of the diffraction limit. 

As a general comment, on the one side, DeepSTORM struggles to provide high precision localization maps: very low Jaccard values for all the tests and tolerances are obtained. For all the tests, the sensitivity reaches the highest possible value but at the expense of a very low specificity, meaning that this result is completely biased since it provides a very huge number of FP molecules (more than 50000). As an example, in Test 2b, this aspect is evident from Fig. \ref{fig:localization} and Fig. \ref{fig:ISBI} where we provide the localization maps for a single frame and for the sum of the stacked SR frames. On the other side, CEL0 reaches the best performances on Test 4b and competing performances on Test2b, but poor results on Test 1b and Test 2b that represent the most realistic and challenging scenarios. Conversely to the competing methods, DeepCEL0 provides satisfying results for all the tests with respect to the Jaccard, sensitivity and specificity indexes thus confirming the effectiveness of the novelties introduced, namely the addition of positivity constraints and the usage of a CEL0 regularized loss function. In particular, the highest value of sensitivity and specificity entails that DeepCEL0 reaches the largest number of TP molecules and reduces the number of FP molecules. Moreover, Fig. \ref{fig:localization} and Fig. \ref{fig:ISBI} confirm once again how our proposal can better provide localization maps with less FP molecules than CEL0 solutions. 

Finally, an interesting aspect to underline is that, even if DeepCEL0 is trained on images with SNR equals 15 dB on average, the trained models seem stable at varying level of corruption (Test 1b, Test 2b, Test 3b). This stability is not still valid for CEL0 which shows a strong dependence on the choice of the regularization parameter as highlighted by the results in Table \ref{tab:1}. 



\input{fig1}

\begin{table*}[!t]

\begin{center}
\caption{\textbf{Performance evaluation for localization on IEEE ISBI simulated dataset in terms of Jaccard Index, sensitivity and specificity.} The first and second best Jaccard Index values are highlighted in red and blue, respectively.} 
\scalebox{1}{
\begin{tabular}{c| c|ccc|c|c}\thickhline 
 \textbf{Test} &  \textbf{Method}   &  \multicolumn{3}{c|}{\textbf{Jaccard } ($\%$)} & \textbf{Sensitivity}($\%$)  & \textbf{Specificity} ($\%$) \\

 &  & $\delta=2$ & $\delta=4$ & $\delta=6$ & $\delta=2$ & $\delta=2$  \\
\thickhline
   \multirow{5}{*}{\textbf{1b}} & \multirow{2}{*}{CEL0} & \textcolor{blue}{41.64}   & \textcolor{blue}{49.46}   & \textcolor{blue}{51.65}   &    56.22  & 99.88    \\
   &                      &  35.94  & 43.93   &  46.70   &  67.74  &  99.71   \\
& DeepSTORM                        &  0.38   & 0.38    & 0.38    &    100.00 & 13.17    \\ 
& DeepCEL0                         & \textcolor{red}{58.96}   & \textcolor{red}{68.94}   & \textcolor{red}{71.55}  &    70.89  & 99.95    
 \\\thickhline 
\multirow{5}{*}{\textbf{2b}} & \multirow{2}{*}{CEL0} & \textcolor{blue}{42.29}   & \textcolor{blue}{50.21}   & \textcolor{blue}{51.48}   &    49.08  & 99.95    \\ 
&                      & 17.13   & 28.64   & 33.16    &    51.15  & 99.34    \\
& DeepSTORM             &  0.38   & 0.38    & 0.38    &    100.00 & 13.34    \\ 
& DeepCEL0              & \textcolor{red}{54.83}   & \textcolor{red}{68.49}   & \textcolor{red}{71.37}   &    65.44  & 99.94    
 \\ \thickhline 
  \multirow{5}{*}{\textbf{3b}} & \multirow{2}{*}{CEL0} & \textcolor{red}{62.17}   & \textcolor{blue}{63.60}   & \textcolor{blue}{64.32}   &    65.90  & 99.98    \\ 
                            &                       & 48.19   & 50.46   & 51.38   &    73.73  & 99.82    \\
& DeepSTORM                 &  0.38   & 0.38    & 0.38    &    100.00 & 13.53    \\ 
& DeepCEL0                  & \textcolor{blue}{61.13}   & \textcolor{red}{69.66}   & \textcolor{red}{71.55}   &    69.59  & 99.95    
 \\\thickhline 
 \multirow{4}{*}{\textbf{4b}} & CEL0 & \textcolor{red}{68.27}   & \textcolor{red}{70.33}   & \textcolor{blue}{71.02}   &    78.34  & 99.95    \\
& DeepSTORM                        &  0.38   & 0.38    & 0.38    &    100.00 & 12.65    \\ 
& DeepCEL0                         & \textcolor{blue}{61.48}   & \textcolor{blue}{69.80}   & \textcolor{red}{72.33}   &    70.12  & 99.96

\end{tabular} \label{tab:1}
}
\end{center}

\end{table*}

\input{fig_loc}
\input{fig3}

\subsection{Localization on IEEE ISBI real dataset}

As final evaluation test, we compare DeepCEL0 with CEL0 and DeepSTORM on a real high density dataset of tubulins that is part of the 2013 IEEE ISBI SMLM challenge. The dataset refers to a real acquisition stack representing a field of view of $128 \times 128$ pixels of size 100 nm over 500 different consecutive frames. The PSF is modelled by using a Gaussian function of FWHM = 351.8 nm, that is  $\sigma_{\textit{k}} \approx$ 150 nm \citep{chahid2014echantillonnage}. We aim at representing the acquired LR frames on a finer pixel grid of a factor $L=4$, thus achieving a final resolution of 25 nm as pixel size. 
CEL0 reconstruction has been obtained by choosing the regularization parameter equal to 0.5 as suggested by \citep{bechensteen2019new}. For all the learning-based models, we trained the networks on a training set of synthetic images (\textit{see} Sub-section 3.3) by a Gaussian blur modelled with the above underlined PSF.
In Fig. \ref{fig:ISBI_real}, we report the normalized LR image, i.e, the sum of all the stacked LR frames. Moreover we depict the reconstructions provided by all the considered competitors. Above the green line we show the image with the activated molecules as white spots, whereas, below the yellow line the normalized images are reported. Finally, three close-ups are considered to better underline the differences between the competing methods. All the considerations highlighted for the simulated scenarios are still valid in this case. We here appreciate the influence of the L0 regularization both in CEL0 and DeepCEL0 reconstructions. Indeed, they look sharper than the solution provided by DeepSTORM. Finally, if compared with CEL0, DeepCEL0 better separates the diverse tubulins and better preserves the continuous shapes.

\input{fig4}

%% file: fig1.tex
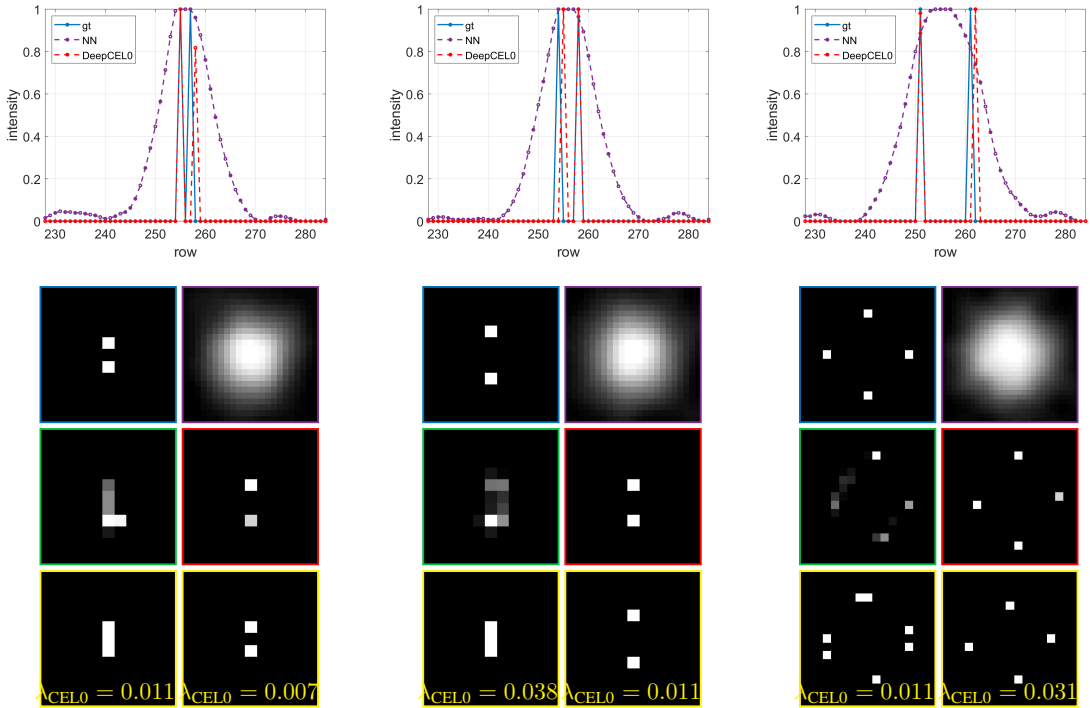
\begin{figure}[H]
	\centering
	\scalebox{0.9}{
	\begin{tikzpicture}
	\begin{scope}[spy using outlines={rectangle,blue_mat,magnification=10,size=2cm}]
	\node {	\includegraphics[height=1cm]{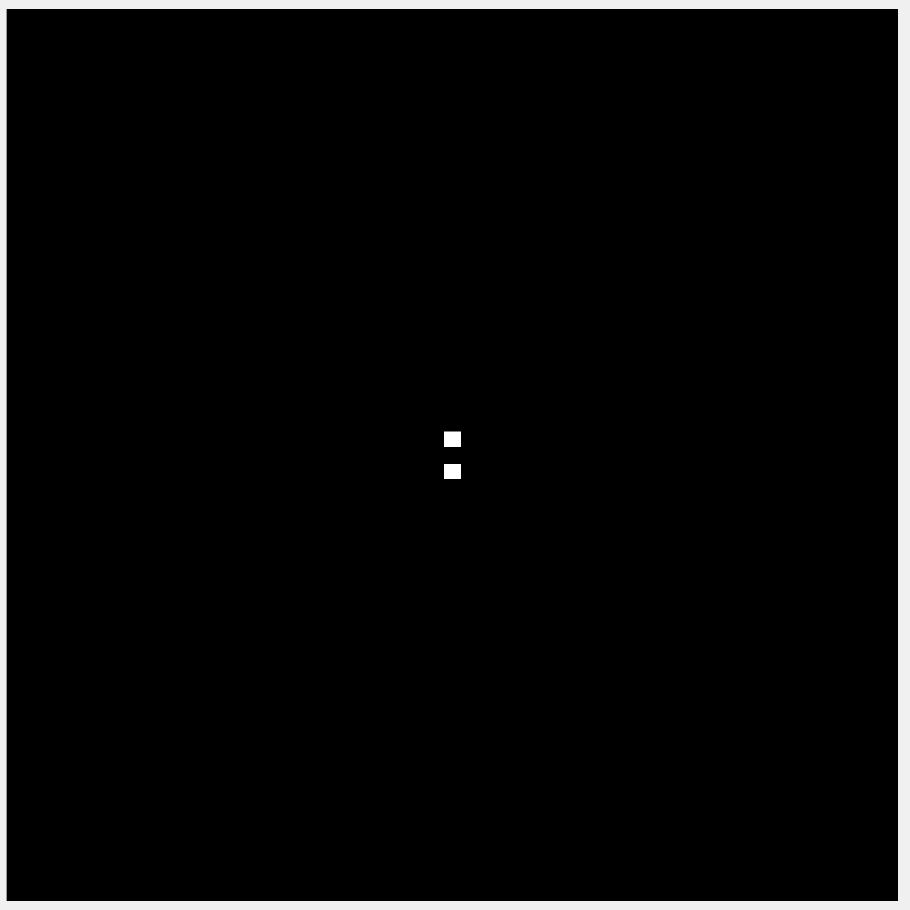}};
	\spy on (-0.0035,0) in node [name=c] at (-1.05,-3.40);
	\end{scope}
	\begin{scope}[spy using outlines={rectangle,purple_mat,magnification=5,size=2cm}]
	\node {	\includegraphics[height=1cm]{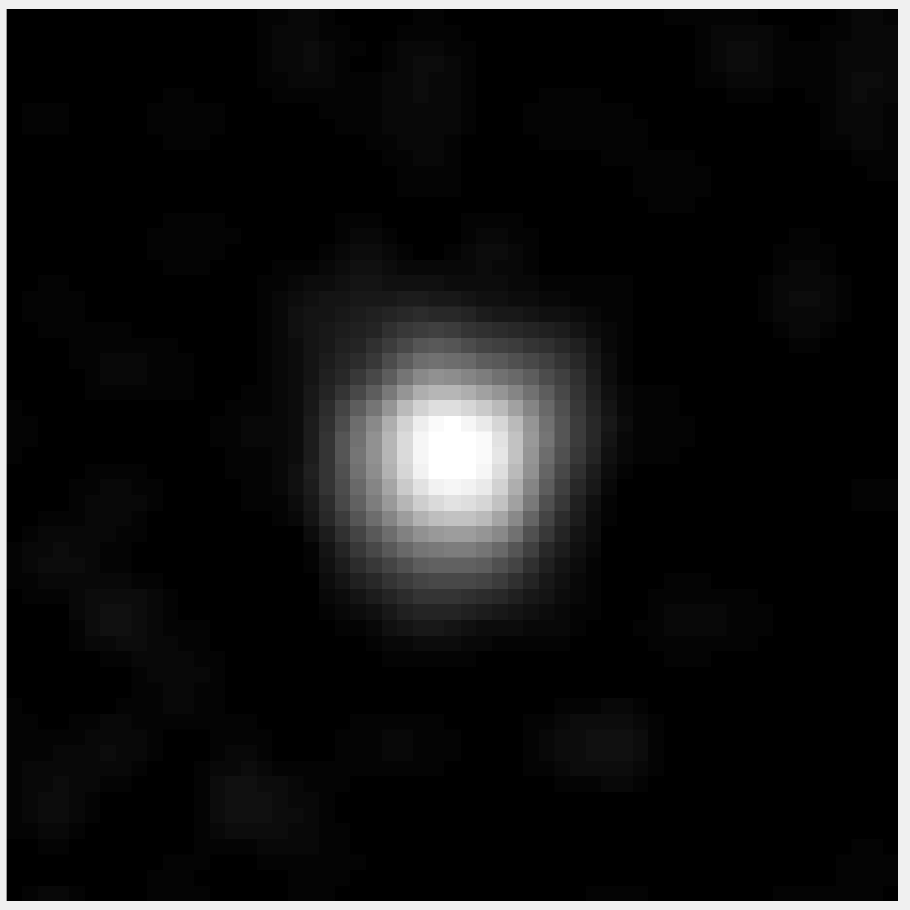}};
	\spy on (-0.0035,0) in node [name=c] at (1.05,-3.40);
    \end{scope}
	\begin{scope}[spy using outlines={rectangle,green_mat,magnification=10,size=2cm}]
	\node {	\includegraphics[height=1cm]{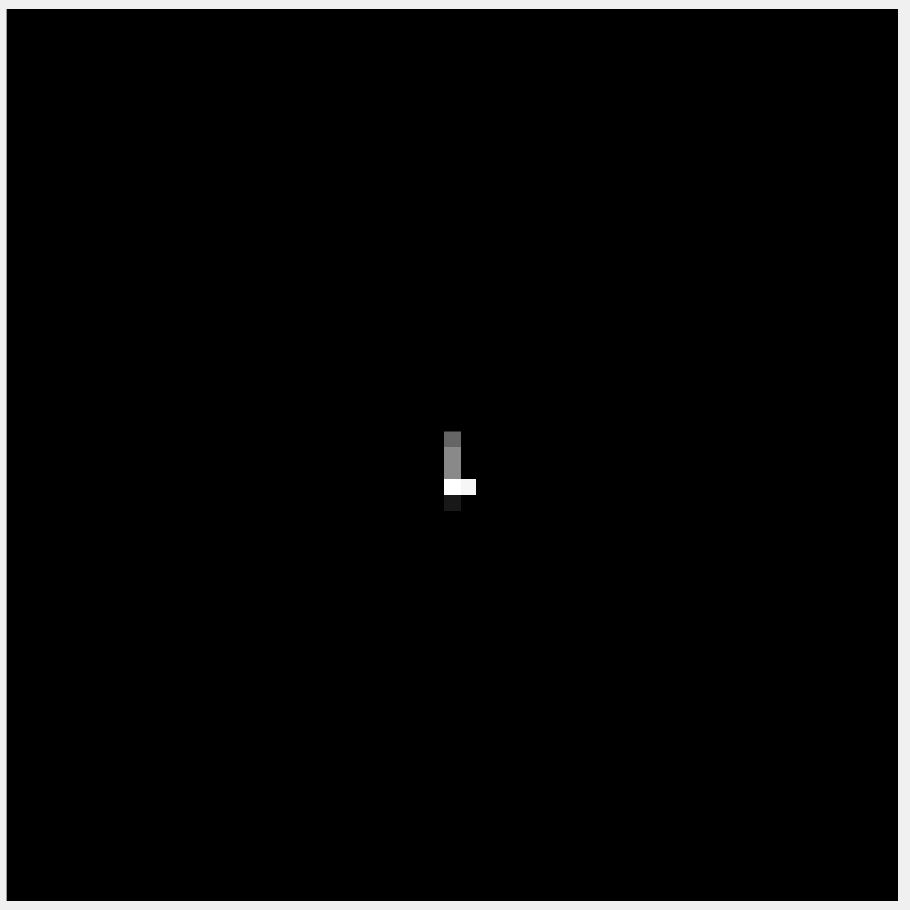}};
	\spy on (-0.0035,0) in node [name=c] at (-1.05,-5.5);
	\end{scope}
	\begin{scope}[spy using outlines={rectangle,red,magnification=10,size=2cm}]
	\node {	\includegraphics[height=1cm]{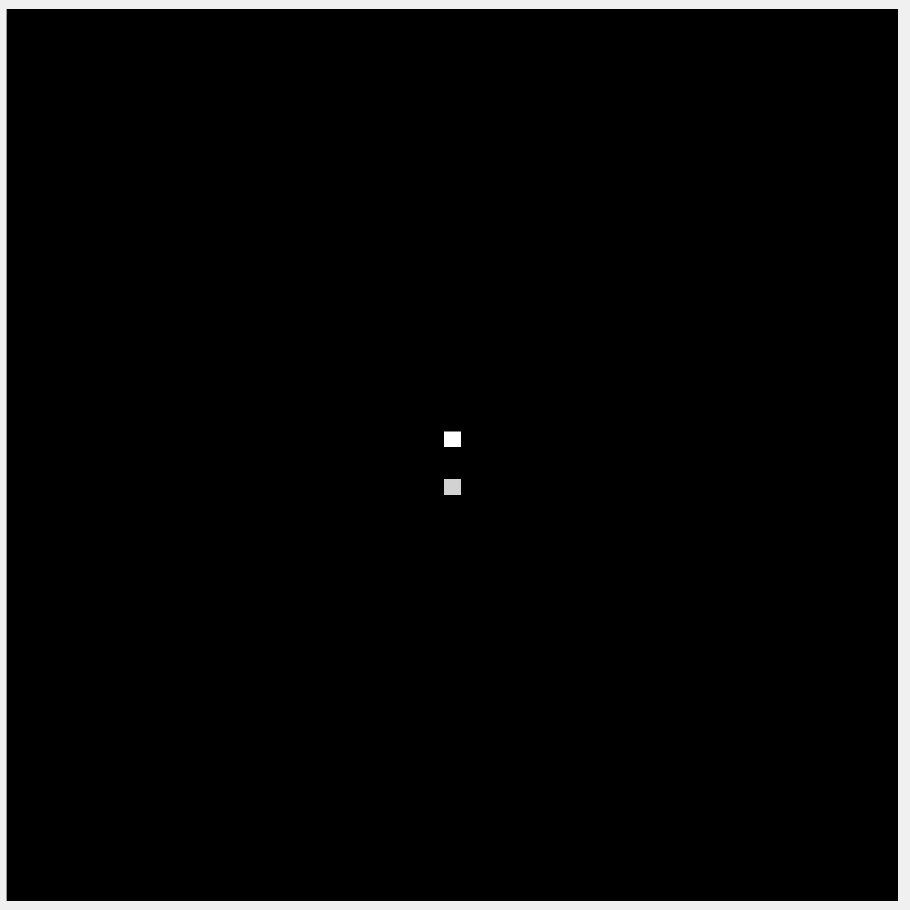}};
	\spy on (-0.0035,0) in node [name=c] at (1.05,-5.5);
	\end{scope}
	\begin{scope}[spy using outlines={rectangle,yellow,magnification=10,size=2cm}]
	\node {	\includegraphics[height=1cm]{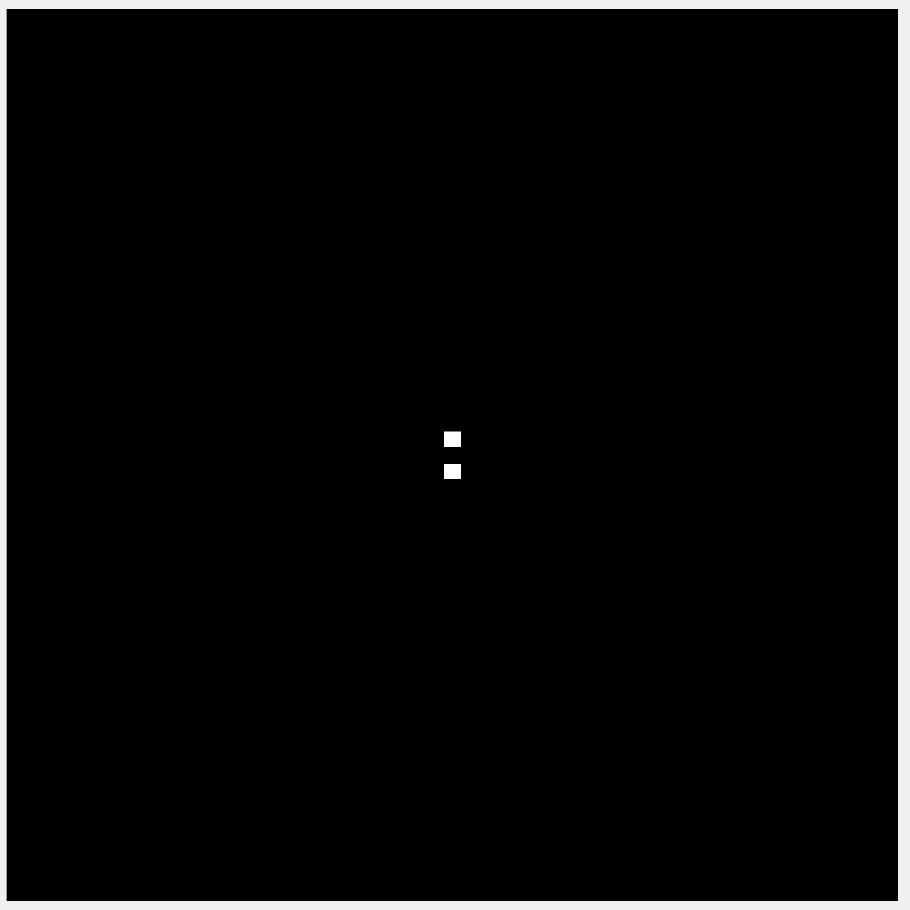}};
	\spy on (-0.0035,0) in node [name=c] at (1.05,-7.6);
	\end{scope}
	\begin{scope}[spy using outlines={rectangle,yellow,magnification=10,size=2cm}]
	\node {	\includegraphics[height=1cm]{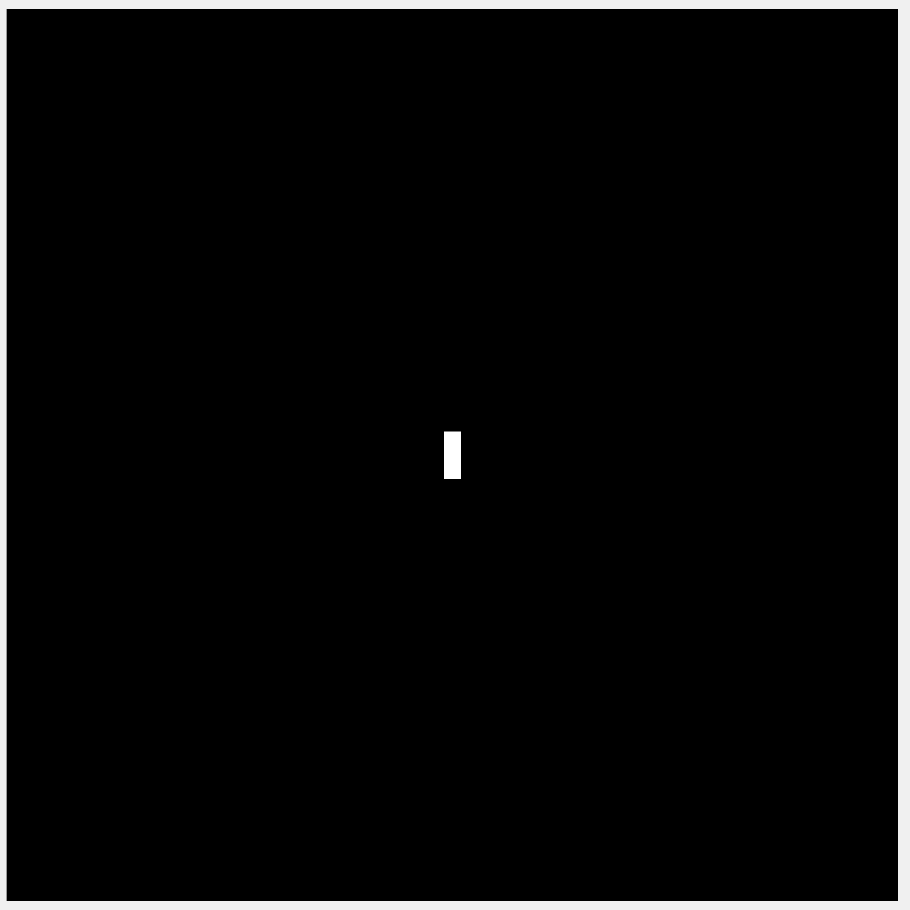}};
	\spy on (-0.0035,0) in node [name=c] at (-1.05,-7.6);
	\end{scope}
	\begin{scope}[spy using outlines={rectangle,black,size=4cm}]
	\node {	\includegraphics[height=4cm]{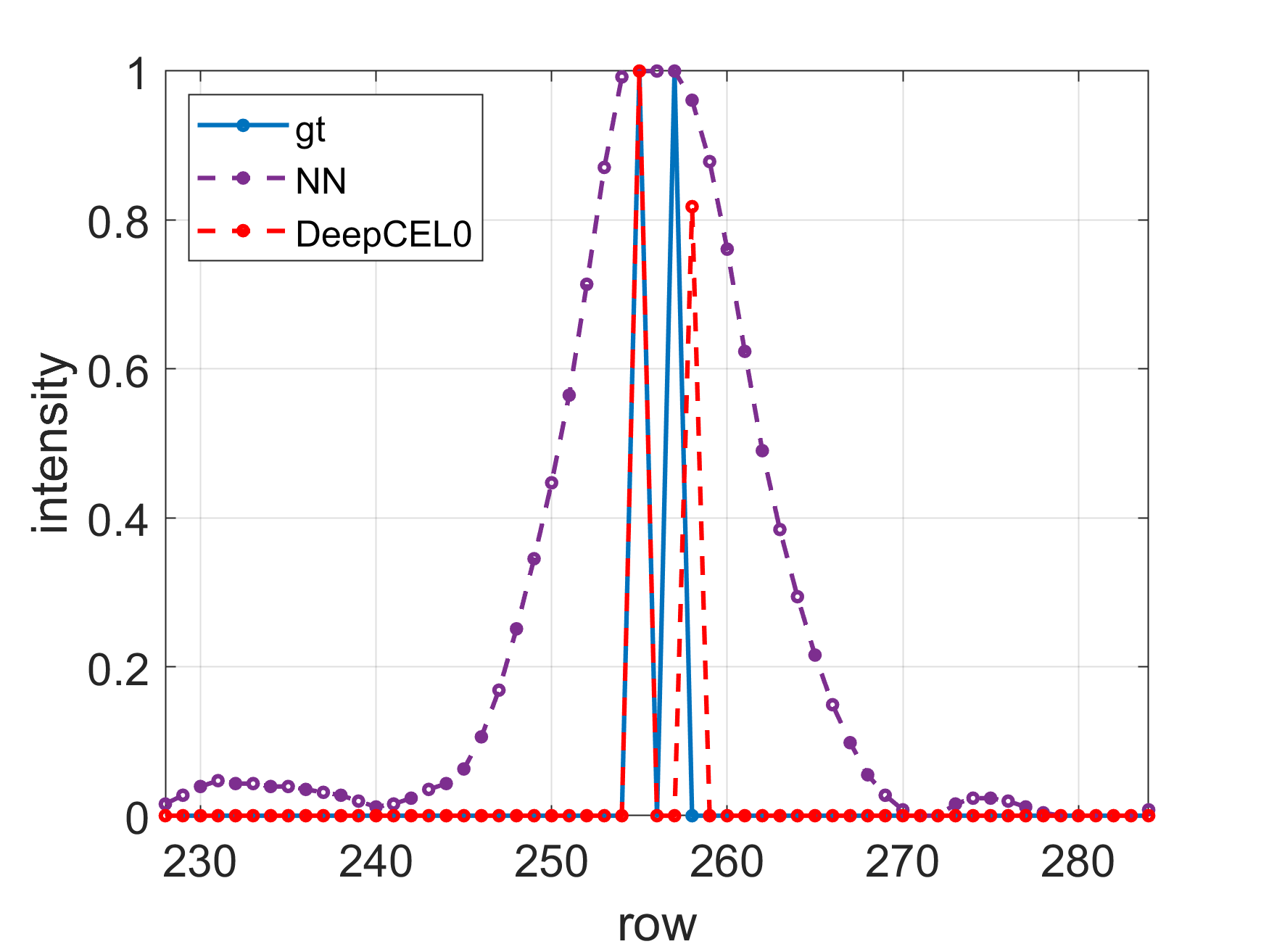}};
	\end{scope}
	\begin{scope}
	\node[yellow] at (-1.1,-8.4) {$\lambda_{\text{CEL0}}=0.011$};
	\end{scope}
	\begin{scope}
	\node[yellow] at (1.02,-8.4) {$\lambda_{\text{CEL0}}=0.007$};
	\end{scope}
	\end{tikzpicture} \begin{tikzpicture}
		\begin{scope}[spy using outlines={rectangle,blue_mat,magnification=10,size=2cm}]
	\node {	\includegraphics[height=1cm]{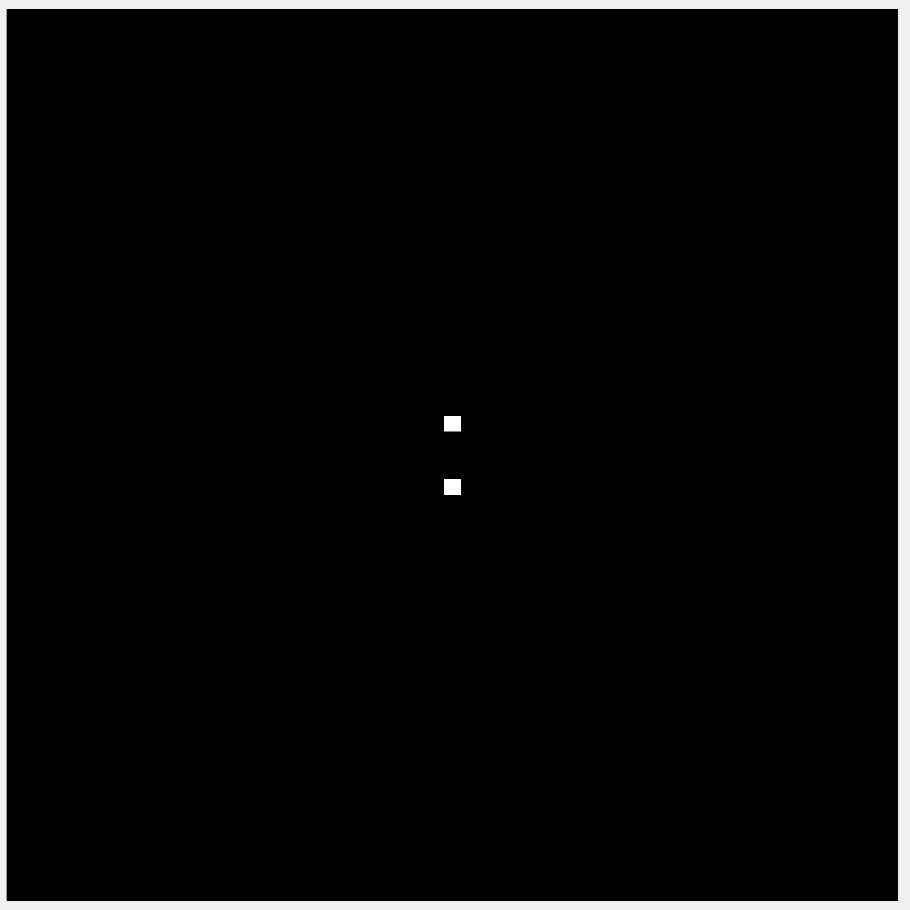}};
	\spy on (-0.0035,0) in node [name=c] at (-1.05,-3.40);
	\end{scope}
	\begin{scope}[spy using outlines={rectangle,purple_mat,magnification=5,size=2cm}]
	\node {	\includegraphics[height=1cm]{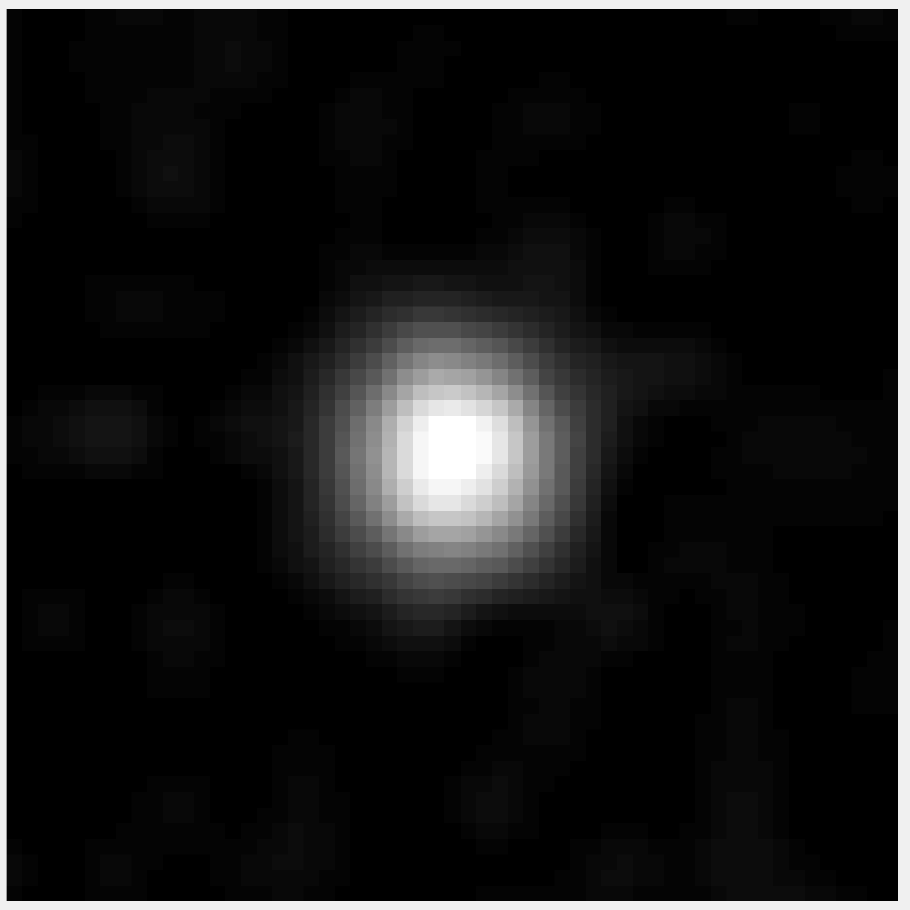}};
	\spy on (-0.0035,0) in node [name=c] at (1.05,-3.40);
    \end{scope}
	\begin{scope}[spy using outlines={rectangle,green_mat,magnification=10,size=2cm}]
	\node {	\includegraphics[height=1cm]{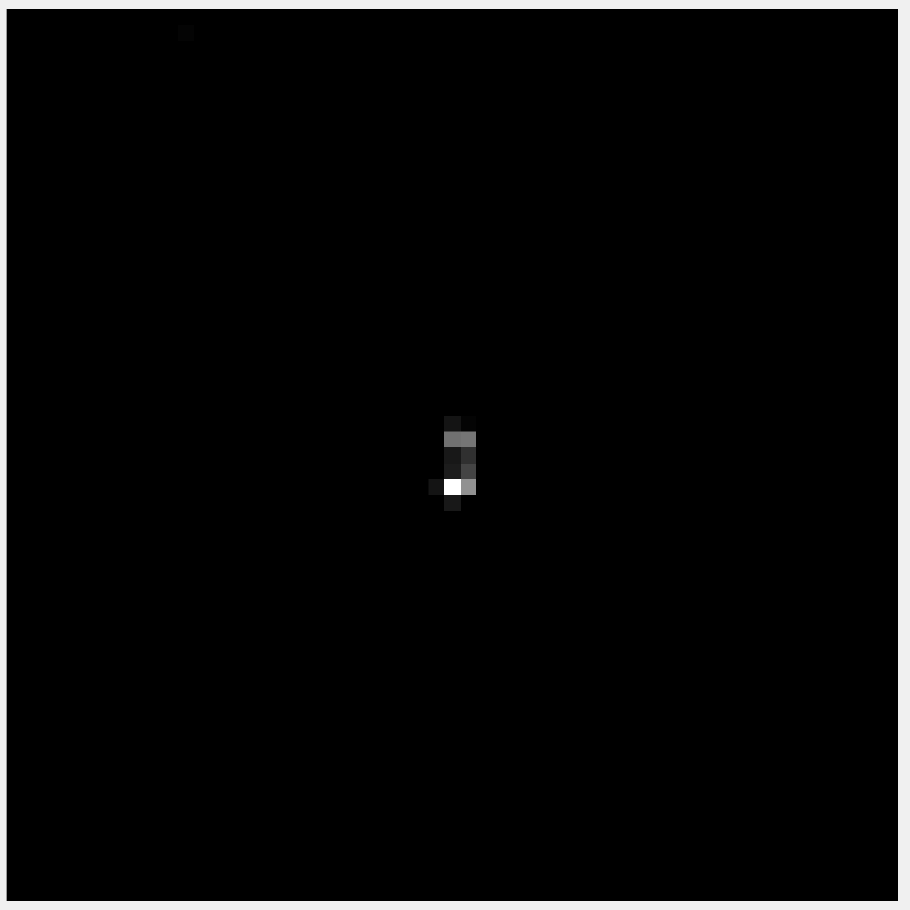}};
	\spy on (-0.0035,0) in node [name=c] at (-1.05,-5.5);
	\end{scope}
	\begin{scope}[spy using outlines={rectangle,red,magnification=10,size=2cm}]
	\node {	\includegraphics[height=1cm]{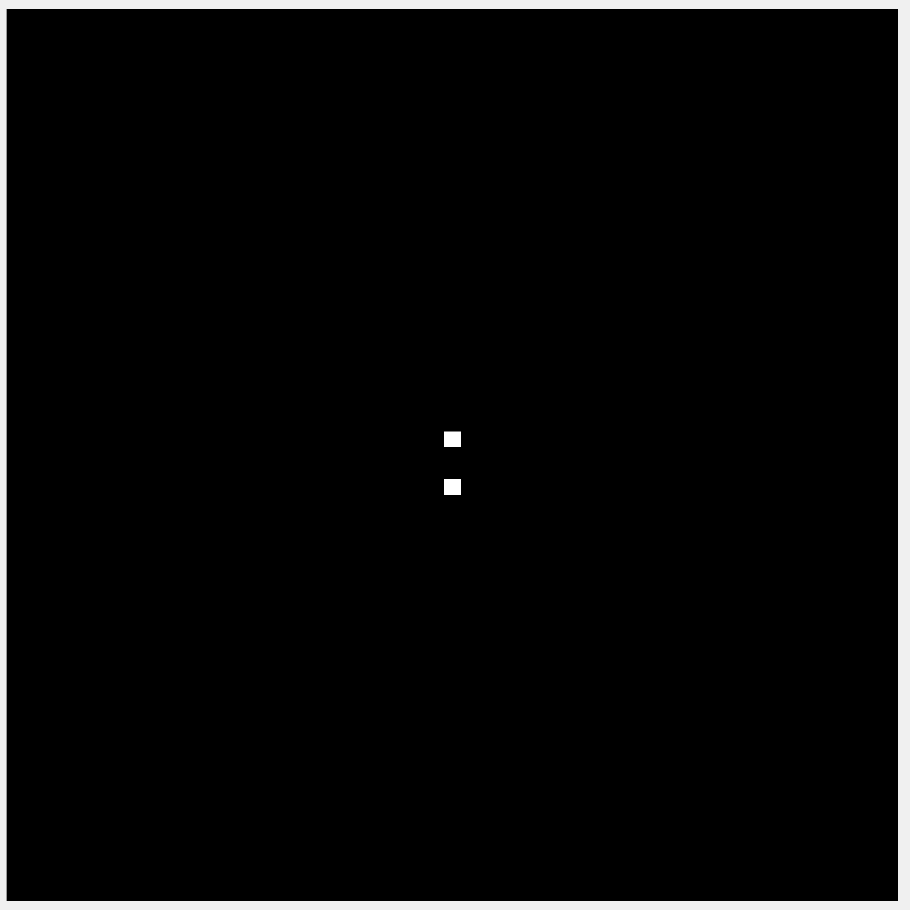}};
	\spy on (-0.0035,0) in node [name=c] at (1.05,-5.5);
	\end{scope}
	\begin{scope}[spy using outlines={rectangle,yellow,magnification=10,size=2cm}]
	\node {	\includegraphics[height=1cm]{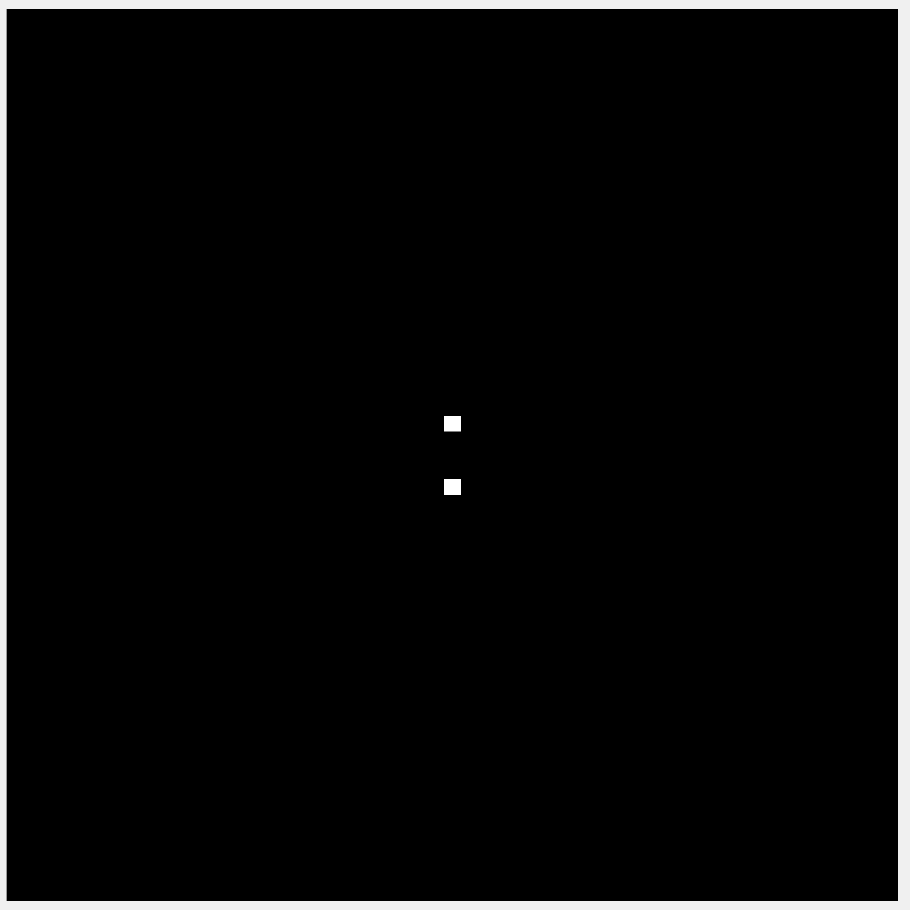}};
	\spy on (-0.0035,0) in node [name=c] at (1.05,-7.6);
	\end{scope}
	\begin{scope}[spy using outlines={rectangle,yellow,magnification=10,size=2cm}]
	\node {	\includegraphics[height=1cm]{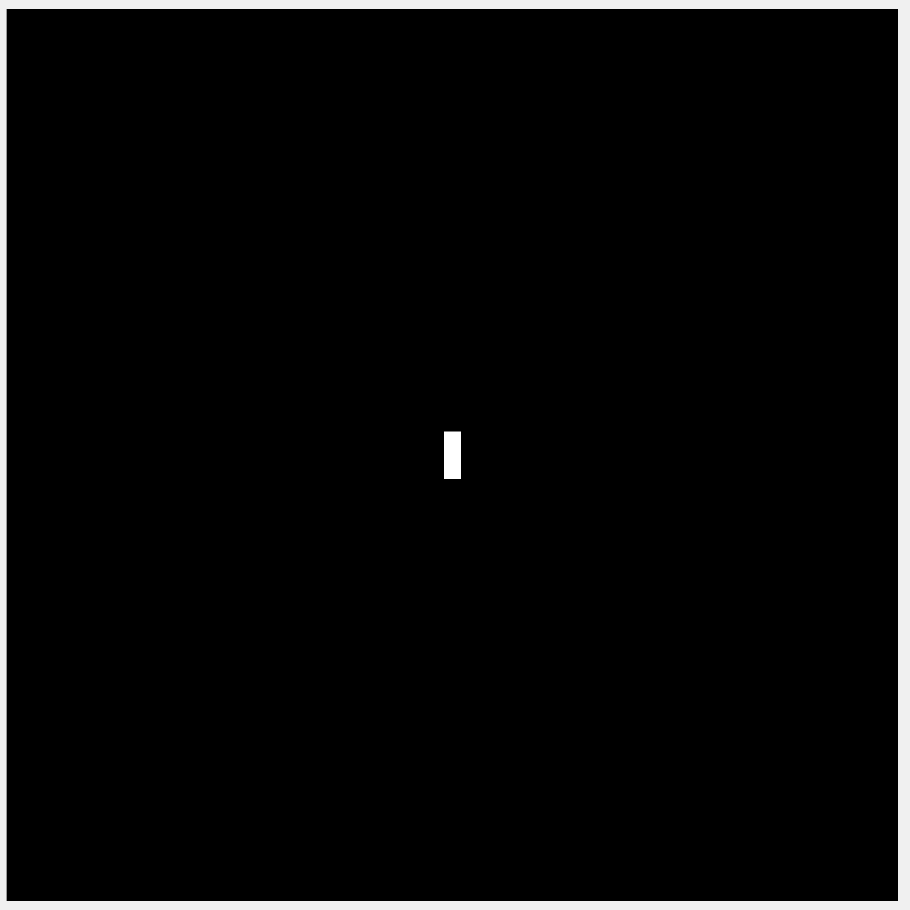}};
	\spy on (-0.0035,0) in node [name=c] at (-1.05,-7.6);
	\end{scope}
	\begin{scope}[spy using outlines={rectangle,black,size=4cm}]
	\node {	\includegraphics[height=4cm]{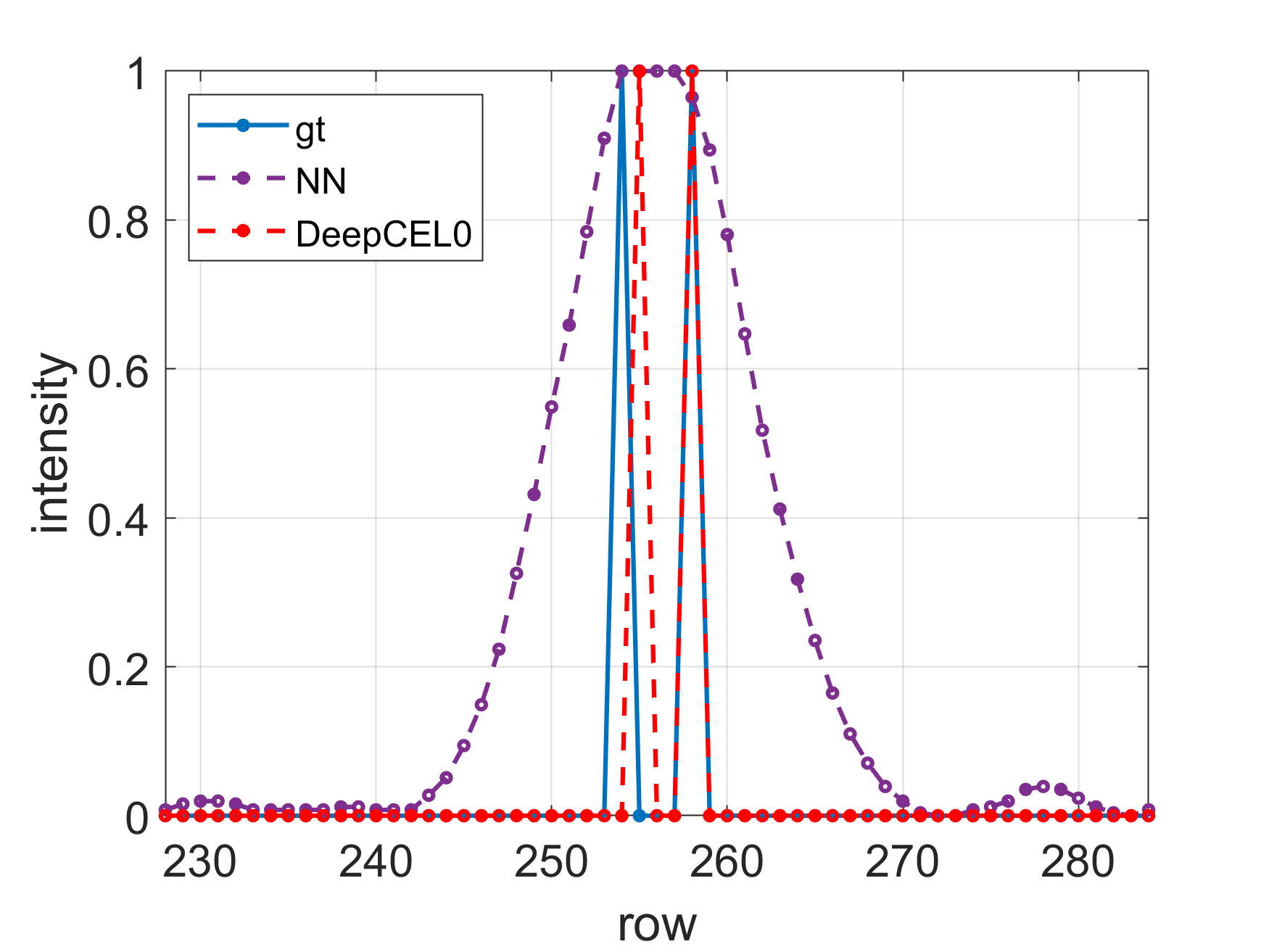}};
	\end{scope}
	\begin{scope}
	\node[yellow] at (-1.1,-8.4) {$\lambda_{\text{CEL0}}=0.038$};
	\end{scope}
	\begin{scope}
	\node[yellow] at (1.02,-8.4) {$\lambda_{\text{CEL0}}=0.011$};
	\end{scope}
	\end{tikzpicture}\begin{tikzpicture}
		\begin{scope}[spy using outlines={rectangle,blue_mat,magnification=7,size=2cm}]
	\node {	\includegraphics[height=1cm]{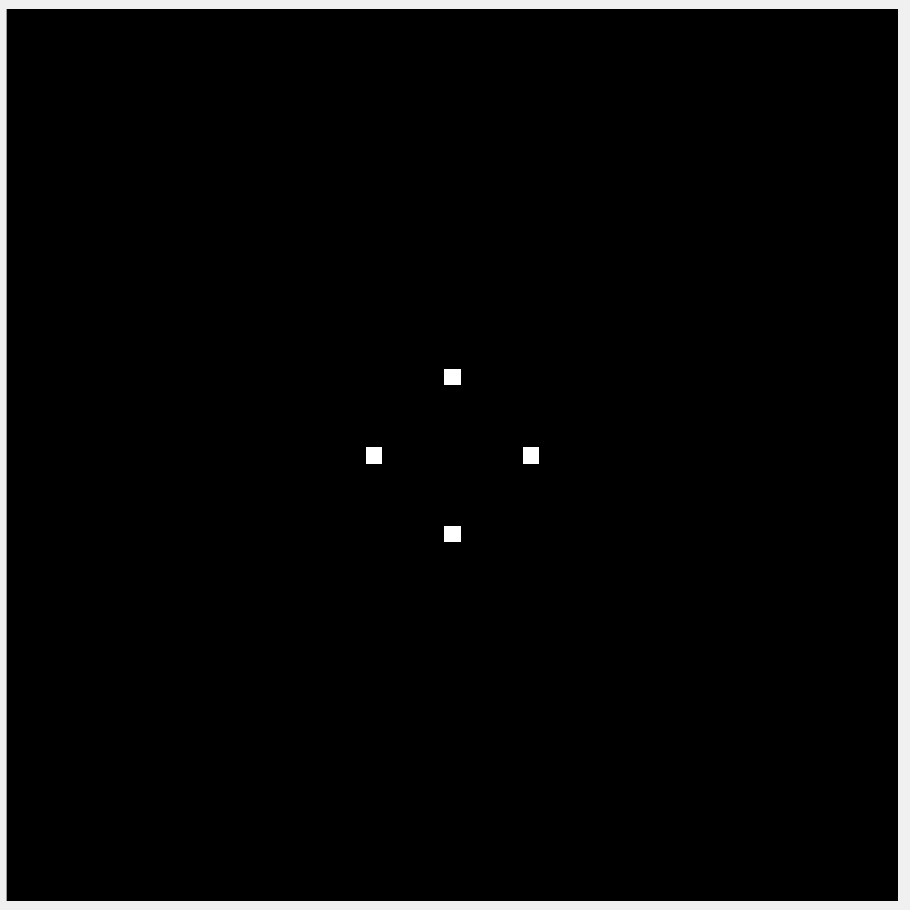}};
	\spy on (-0.0035,0) in node [name=c] at (-1.05,-3.40);
	\end{scope}
	\begin{scope}[spy using outlines={rectangle,purple_mat,magnification=4,size=2cm}]
	\node {	\includegraphics[height=1cm]{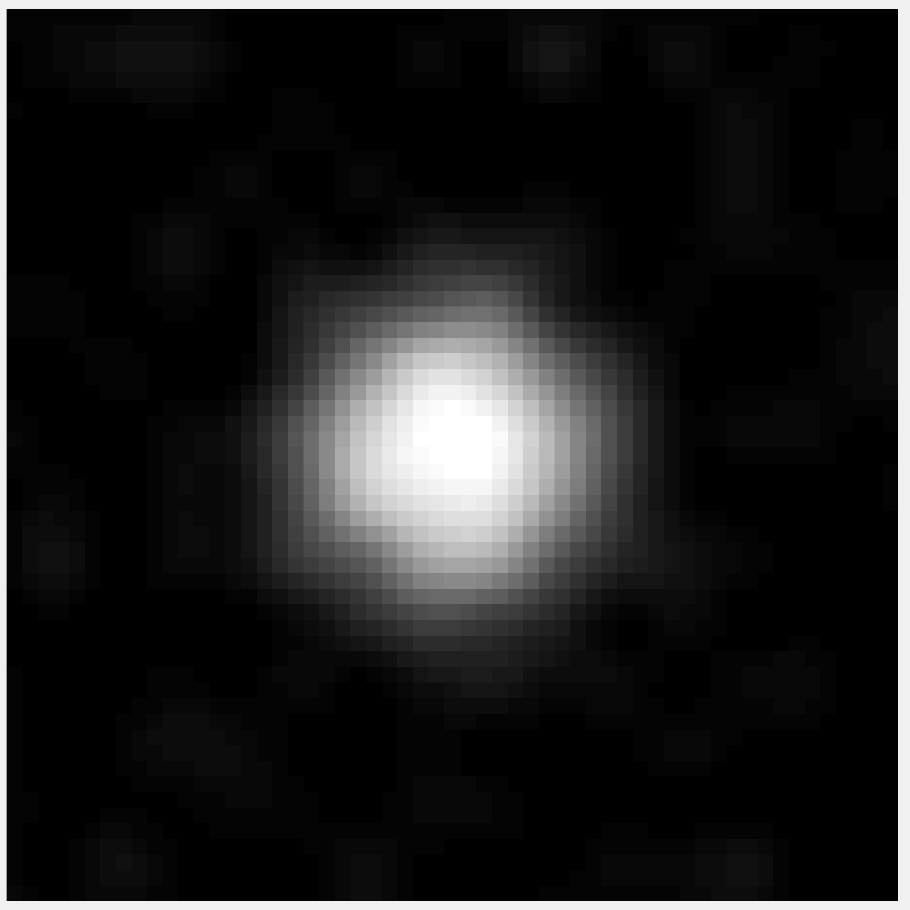}};
	\spy on (-0.0035,0) in node [name=c] at (1.05,-3.40);
    \end{scope}
	\begin{scope}[spy using outlines={rectangle,green_mat,magnification=7,size=2cm}]
	\node {	\includegraphics[height=1cm]{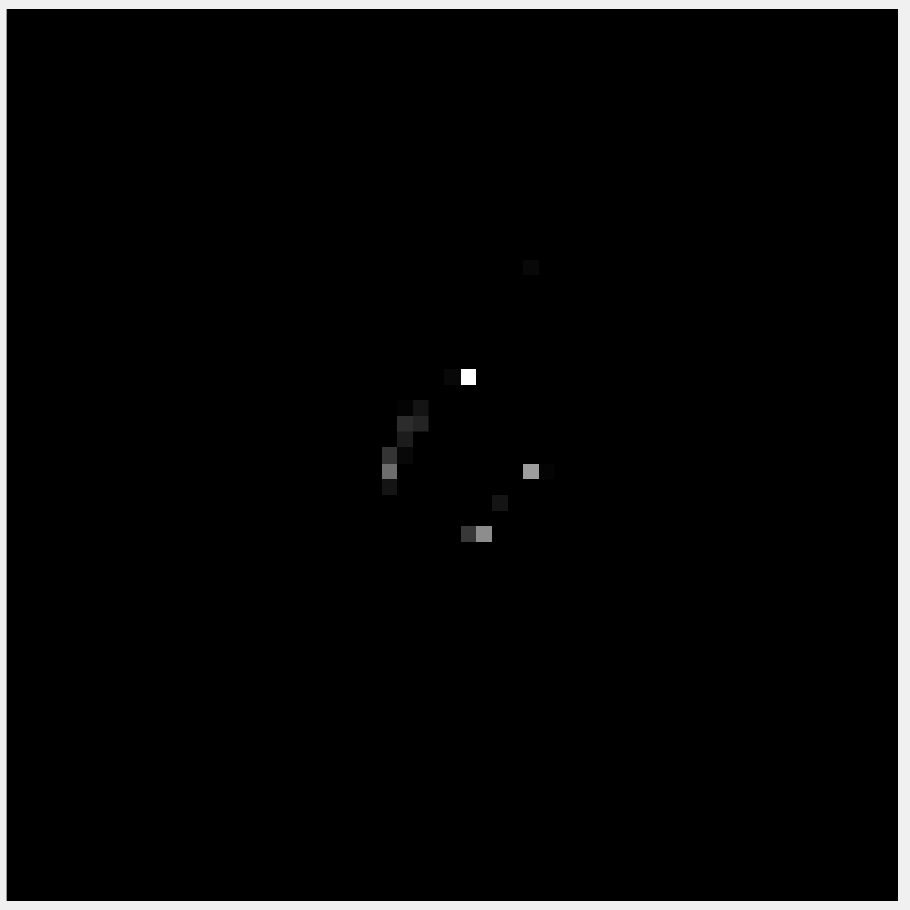}};
	\spy on (-0.0035,0) in node [name=c] at (-1.05,-5.5);
	\end{scope}
	\begin{scope}[spy using outlines={rectangle,red,magnification=7,size=2cm}]
	\node {	\includegraphics[height=1cm]{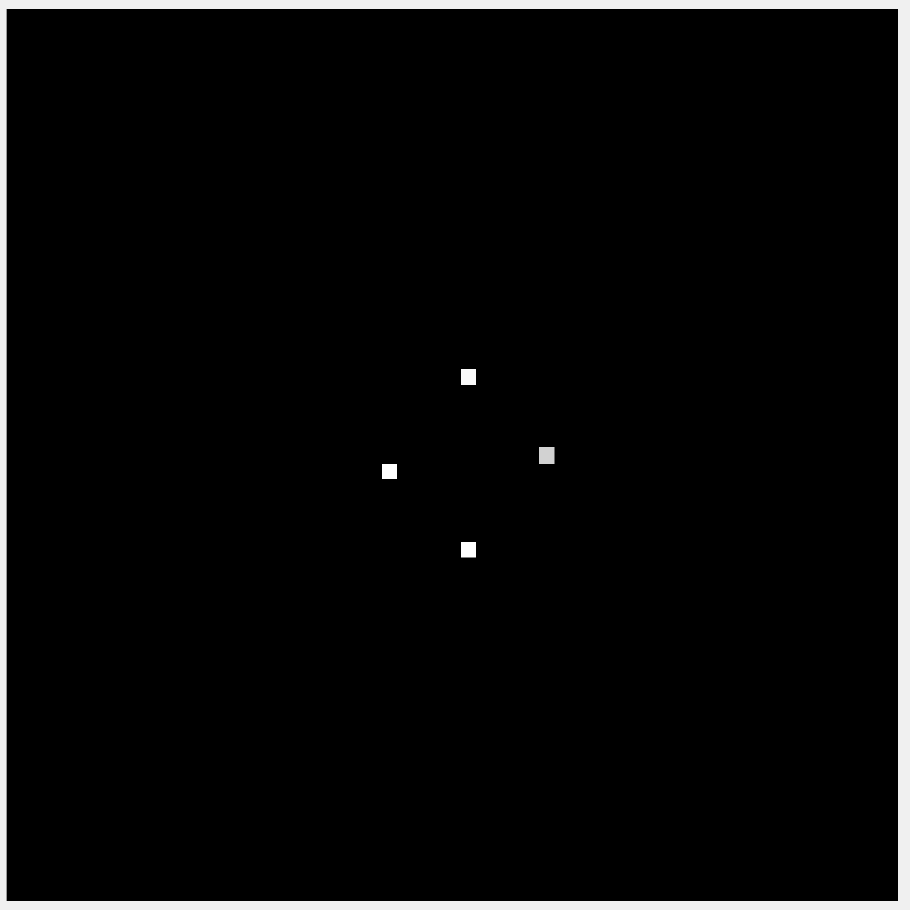}};
	\spy on (-0.0035,0) in node [name=c] at (1.05,-5.5);
	\end{scope}
	\begin{scope}[spy using outlines={rectangle,yellow,magnification=7,size=2cm}]
	\node {	\includegraphics[height=1cm]{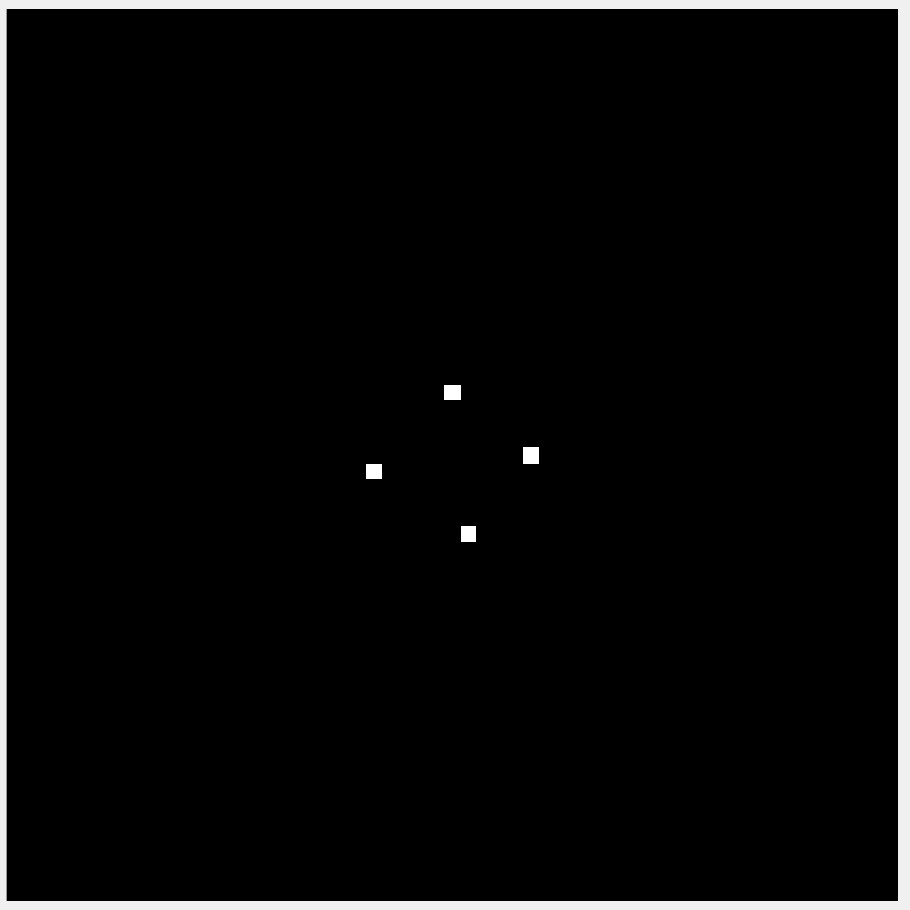}};
	\spy on (-0.0035,0) in node [name=c] at (1.05,-7.6);
	\end{scope}
	\begin{scope}[spy using outlines={rectangle,yellow,magnification=7,size=2cm}]
	\node {	\includegraphics[height=1cm]{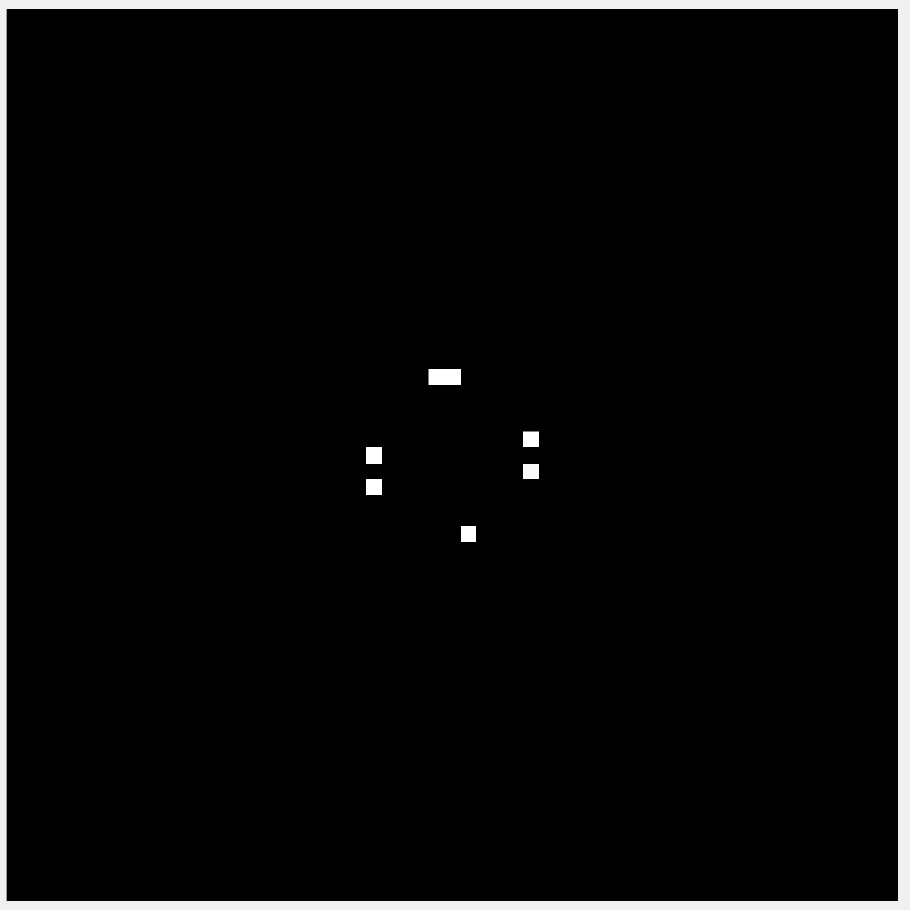}};
	\spy on (-0.0035,0) in node [name=c] at (-1.05,-7.6);
	\end{scope}
	\begin{scope}[spy using outlines={rectangle,black,size=4cm}]
	\node {	\includegraphics[height=4cm]{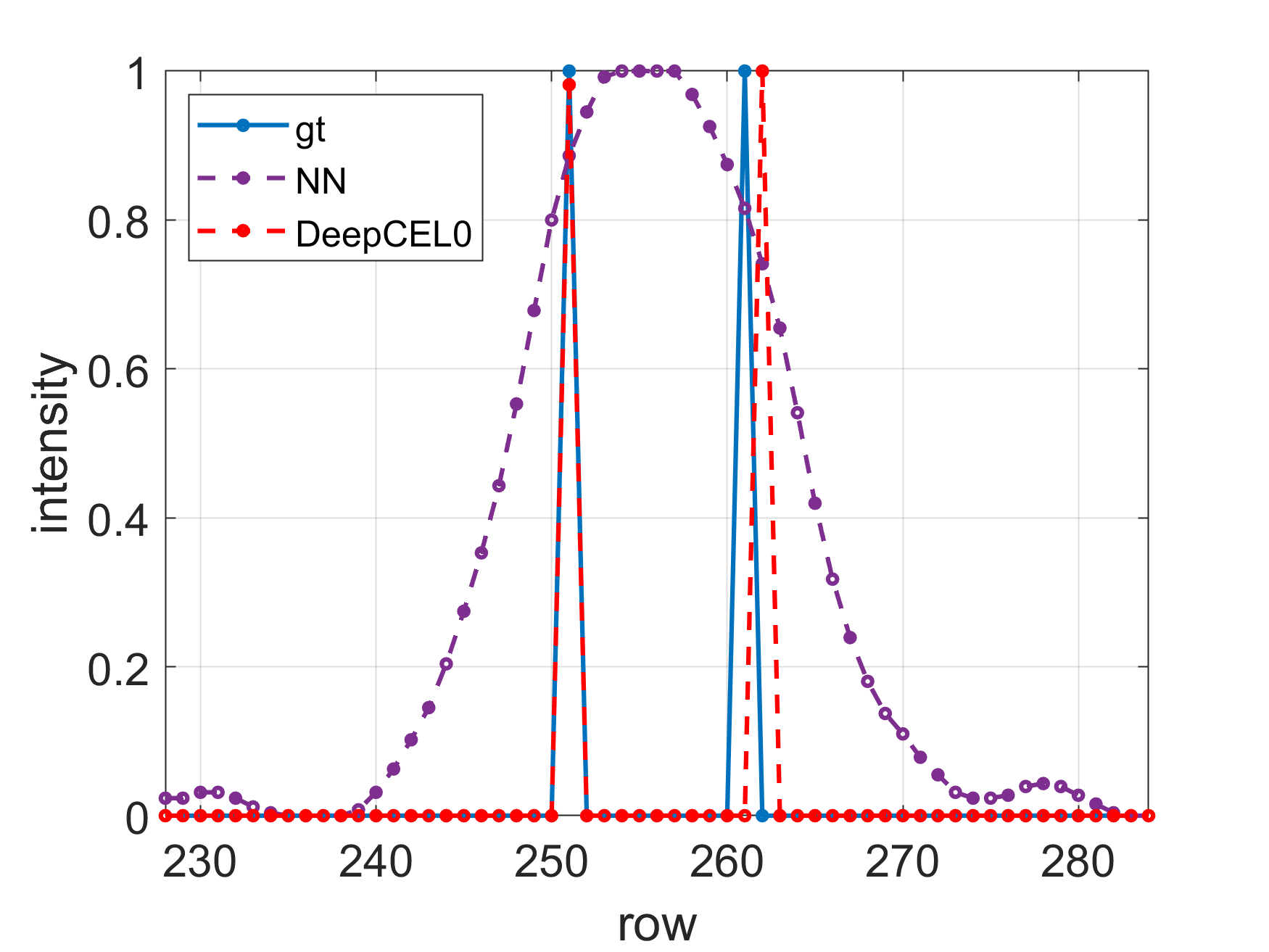}};
	\end{scope}
	\begin{scope}
	\node[yellow] at (-1.1,-8.4) {$\lambda_{\text{CEL0}}=0.011$};
	\end{scope}
	\begin{scope}
	\node[yellow] at (1.02,-8.4) {$\lambda_{\text{CEL0}}=0.031$};
	\end{scope}
	\end{tikzpicture}
	}
	\caption{\textbf{Localization on synthetic test images.} Line profiles crossing the 256th column of GT, NN and DeepCEL0 images for Test 1a, Test 2a, Test 3a (upper panel from left to right). Close-ups (x10) on the ROI of GT (blue box), NN (purple box), DeepSTORM (green box), DeepCEL0 (red box) and CEL0 with two different regularization parameters (yellow box) for Test 1a, Test 2a, Test 3a (lower panel from left to right). }
	\label{fig:sintetic_reconstruction}
	
\end{figure}


%% file: fig_loc.tex
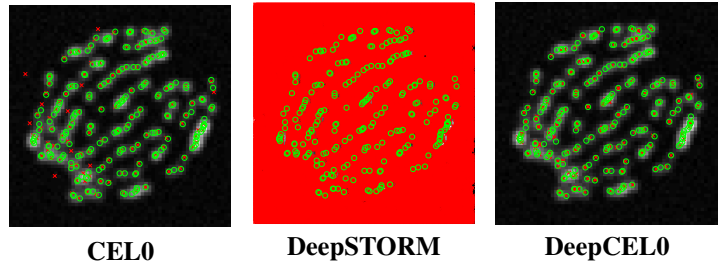
\begin{figure}[H]
	\centering
	
	\begin{tikzpicture}
	\begin{scope}[spy using outlines={rectangle,black,magnification=2,size=2cm}]
	\node[name=c] {	\includegraphics[height=3cm]{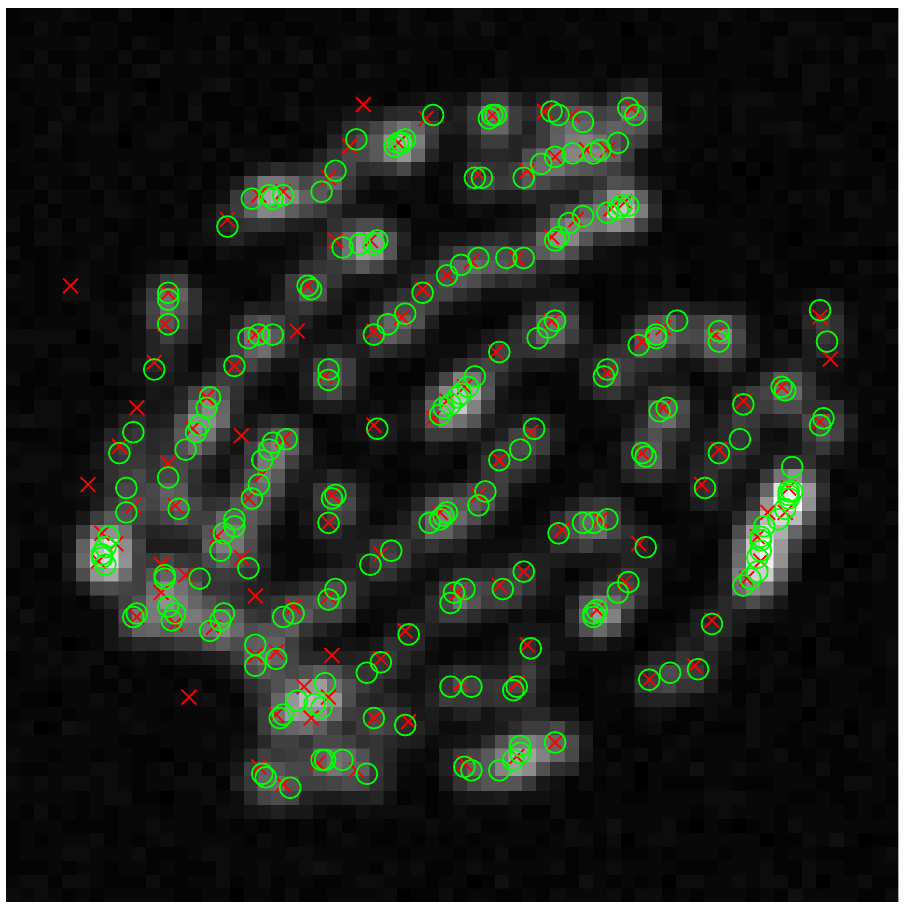}};
	\end{scope}
	\node[below of=c, align=center, node distance=1.8cm] {\textbf{CEL0}};
	\end{tikzpicture}\begin{tikzpicture}
	\begin{scope}[spy using outlines={rectangle,black,magnification=2,size=2cm}]
	\node [name=c]{	\includegraphics[height=3cm]{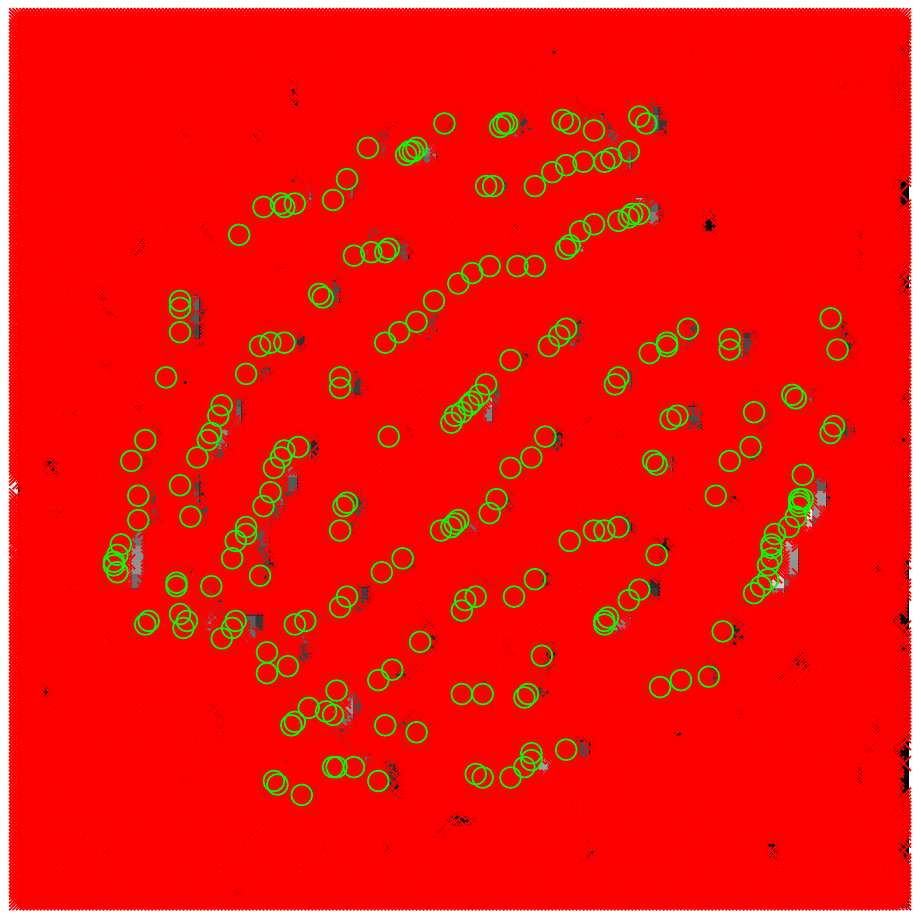}};
	\end{scope}
	\node[below of=c, align=center, node distance=1.8cm] {\textbf{DeepSTORM}};
	\end{tikzpicture}\begin{tikzpicture}
	\begin{scope}[spy using outlines={rectangle,black,magnification=2,size=2cm}]
	\node [name=c]{	\includegraphics[height=3cm]{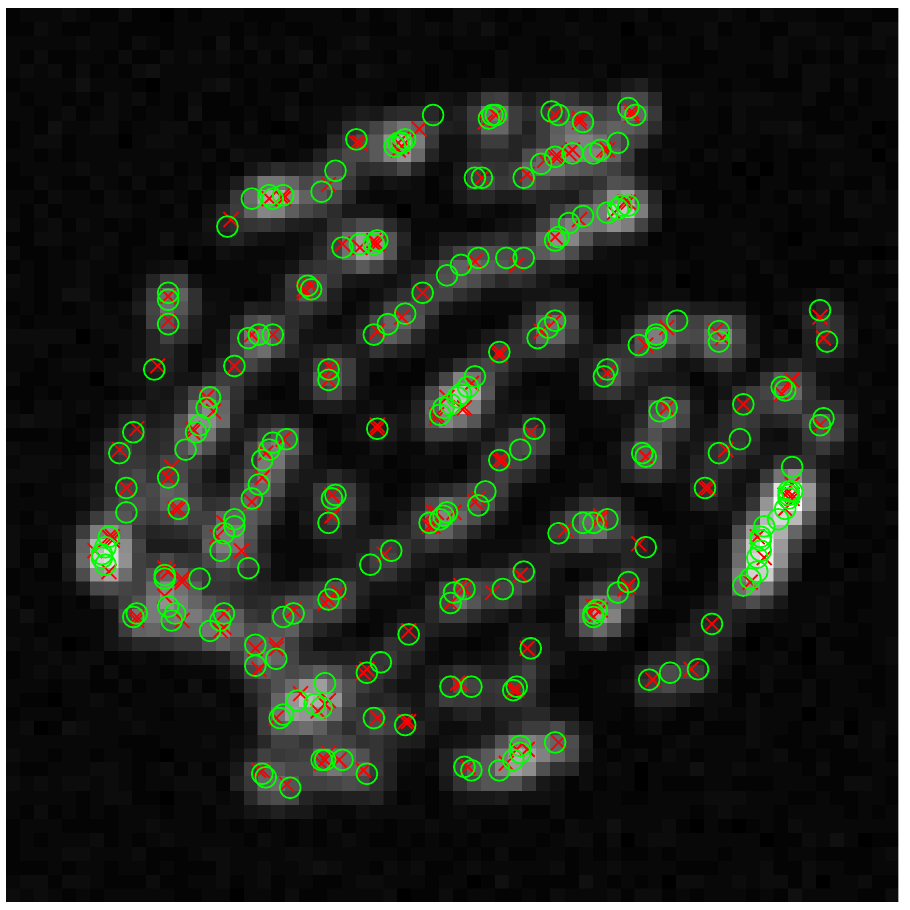}};
	\end{scope}
	\node[below of=c, align=center, node distance=1.8cm] {\textbf{DeepCEL0}};
	\end{tikzpicture}

	\caption{\textbf{Localization maps for the second frame of Test 1b.} The green circles indicate the TP molecules; the red cross marks indicate the FP molecules.}
	\label{fig:localization}
\end{figure}

%% file: fig3.tex
\begin{figure}[H]
    \newlength{\imagewidth}
    \settowidth{\imagewidth}{\includegraphics{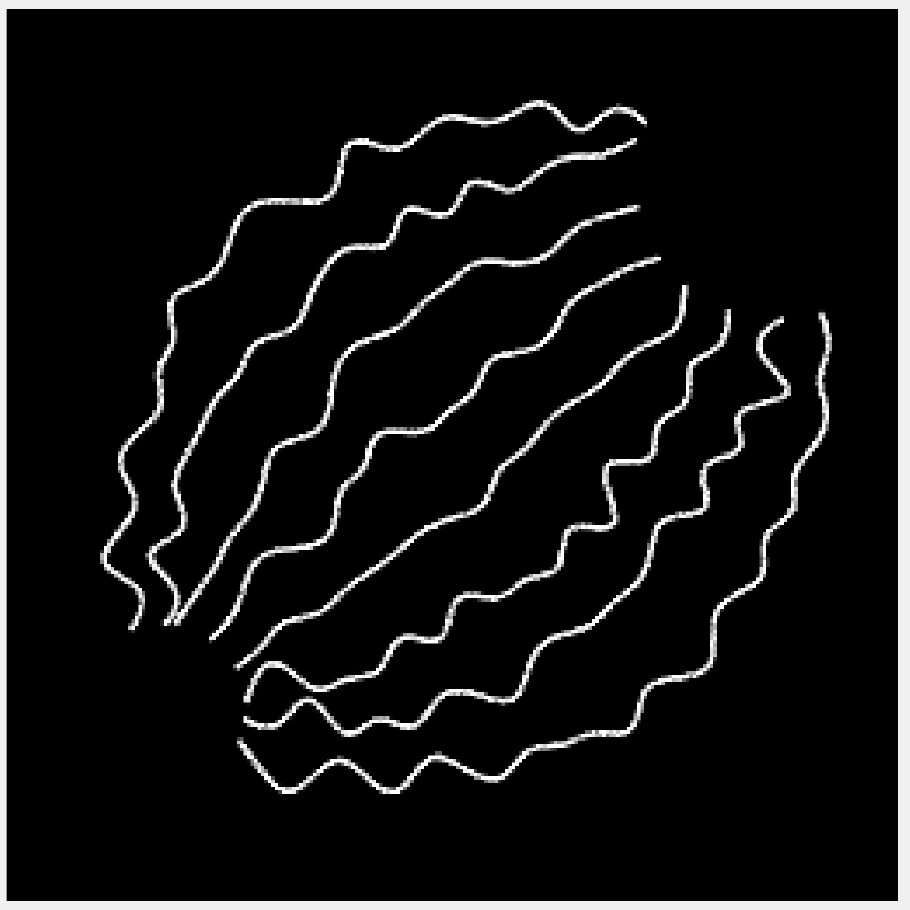}}
	\centering
	\scalebox{1.3}{\begin{tikzpicture}
	\begin{scope}[spy using outlines={circle,black,magnification=4,size=1cm,connect spies}]
	\node [name=c]{	\includegraphics[height=2cm]{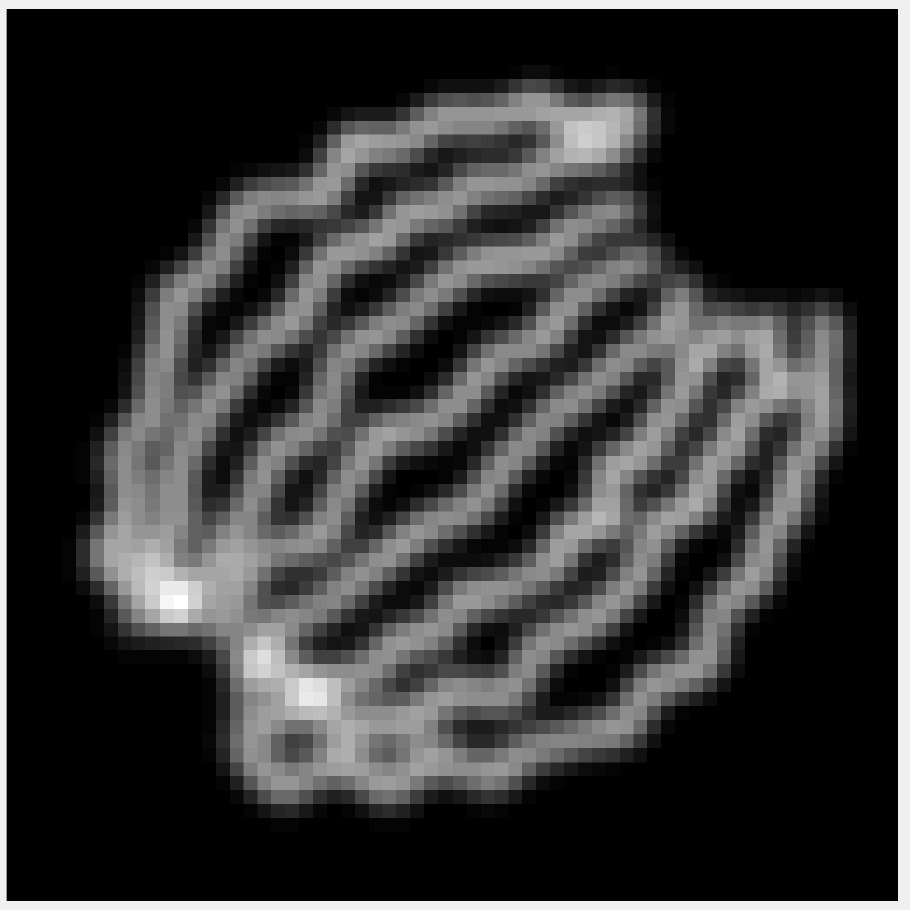}};
	\spy[cyan] on (0.32,0.70) in node [name=d1] at (0.52,1.5);
	\spy[cyan] on (-0.22,-0.65) in node [name=d1] at (0.52,-1.5);
	\spy[cyan] on (-0.65,0) in node [name=d1] at (-0.52,-1.5);
	\end{scope}
	\node[below of=c, align=center, node distance=2.5cm] {\textbf{LR}};
	\end{tikzpicture}\begin{tikzpicture}
	\begin{scope}[spy using outlines={circle,black,magnification=4,size=1cm,connect spies}]
	\node[name=d] at (0,0.5) {\includegraphics[trim=0 0.5\imagewidth{} 0 0,width=2cm, clip]{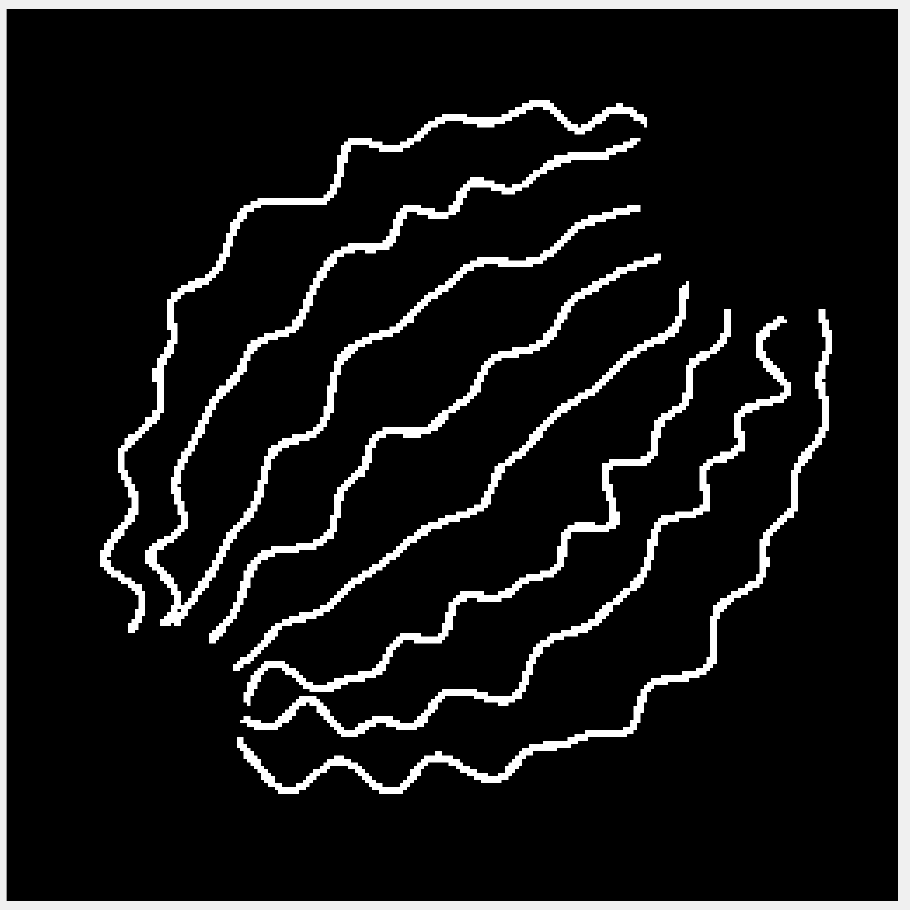}};
	\node[name=b] at (0,-0.5) {\includegraphics[trim=0 0 0 0.5\imagewidth{},width=2cm, clip]{img/ISBI/GT.pdf}};
	\node[name=c] at (0,0) {};
	\spy[cyan] on (0.32,0.70) in node [name=d1] at (0.52,1.5);
	\spy[cyan] on (-0.22,-0.65) in node [name=d1] at (0.52,-1.5);
	\spy[cyan] on (-0.65,0) in node [name=d1] at (-0.52,-1.5);
	\draw[magenta, ultra thin] (-1,0) -- (1,0);
	\end{scope}
	\node[below of=c, align=center, node distance=2.5cm] {\textbf{GT}};
	\end{tikzpicture}\begin{tikzpicture}
	\begin{scope}[spy using outlines={circle,black,magnification=4,size=1cm,connect spies}]
	\node[name=d] at (0,0.5) {\includegraphics[trim=0 0.5\imagewidth{} 0 0,width=2cm, clip]{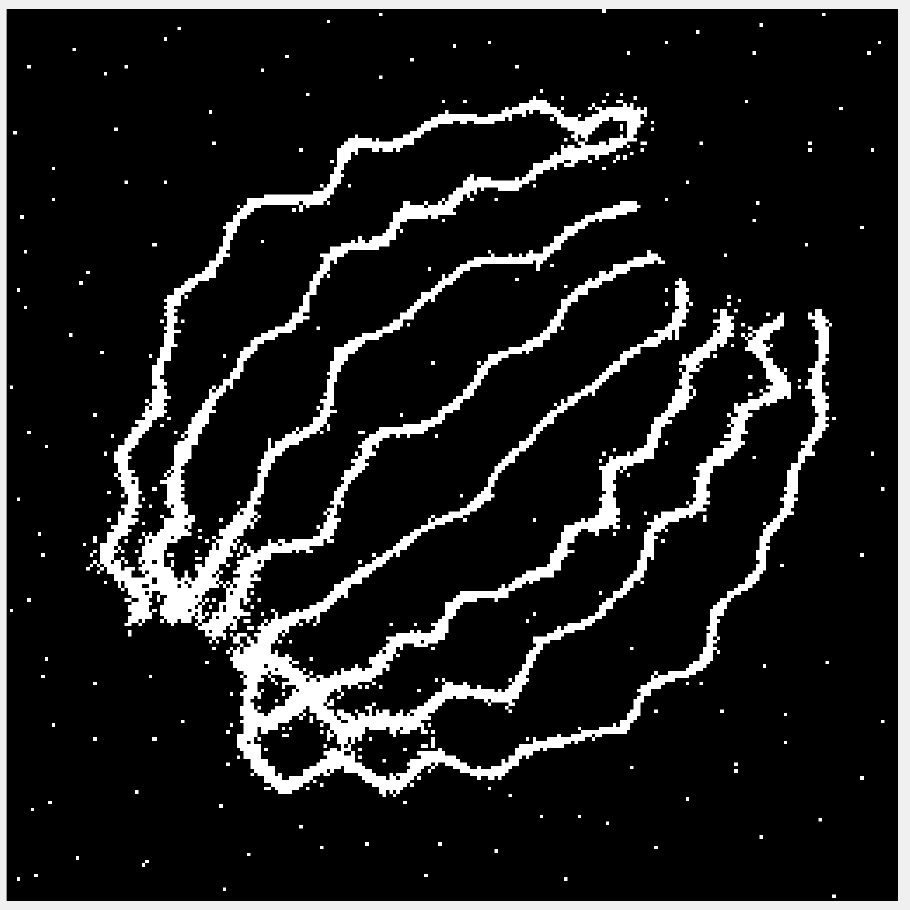}};
	\node[name=b] at (0,-0.5) {\includegraphics[trim=0 0 0 0.5\imagewidth{},width=2cm, clip]{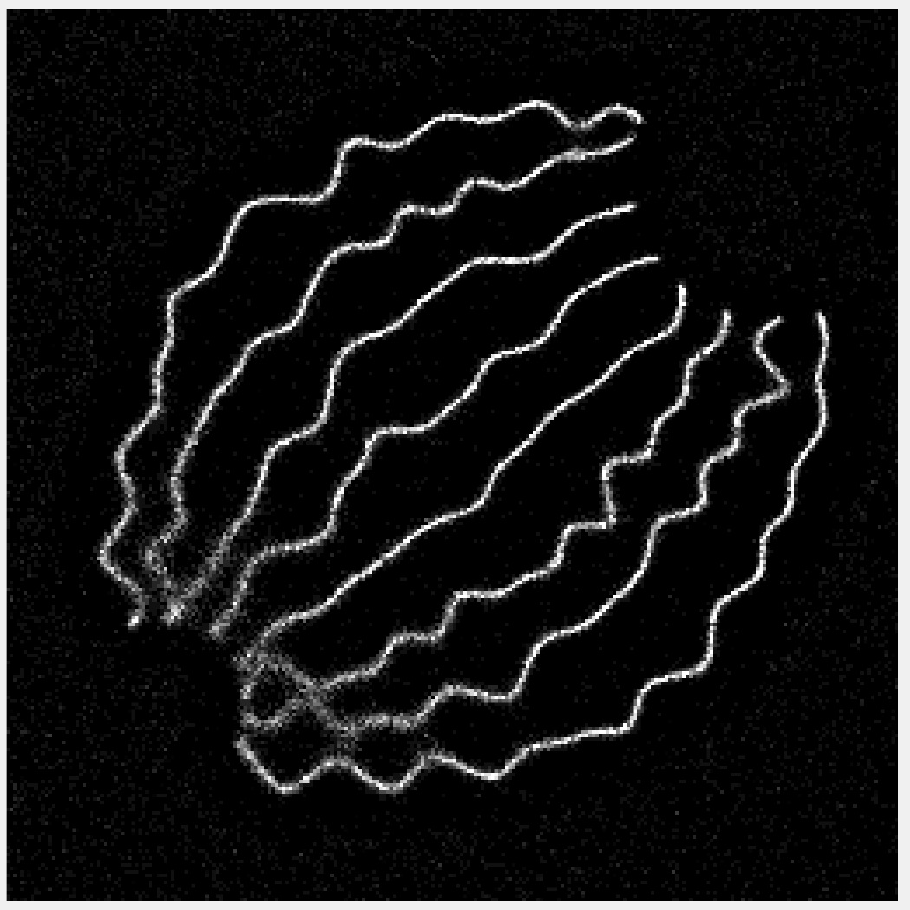}};
	\node[name=c] at (0,0) {};
	\spy[cyan] on (0.32,0.70) in node [name=d1] at (0.52,1.5);
	\spy[cyan] on (-0.22,-0.65) in node [name=d1] at (0.52,-1.5);
	\spy[cyan] on (-0.65,0) in node [name=d1] at (-0.52,-1.5);
	\draw[magenta, ultra thin] (-1,0) -- (1,0);
	\end{scope}
	\node[below of=c, align=center, node distance=2.5cm] {\textbf{CEL0}};
	\end{tikzpicture}\begin{tikzpicture}
	\begin{scope}[spy using outlines={circle,black,magnification=4,size=1cm,connect spies}]
	\node[name=d] at (0,0.5) {\includegraphics[trim=0 0.5\imagewidth{} 0 0,width=2cm, clip]{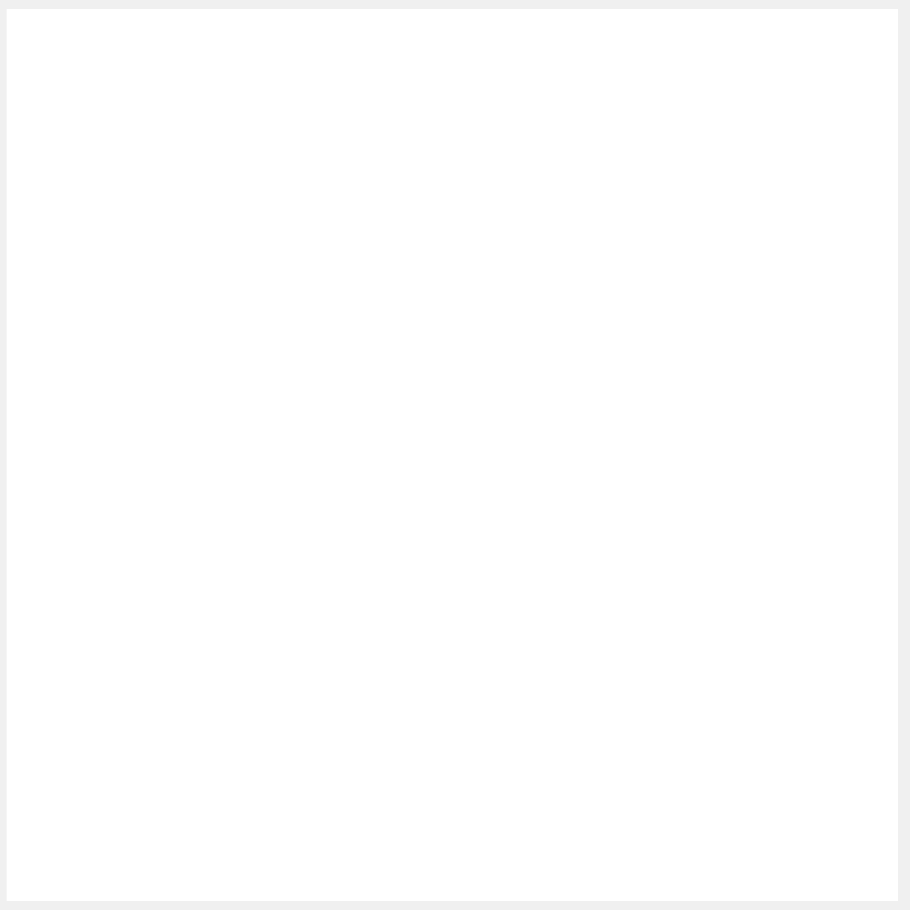}};
	\node[name=b] at (0,-0.5) {\includegraphics[trim=0 0 0 0.5\imagewidth{},width=2cm, clip]{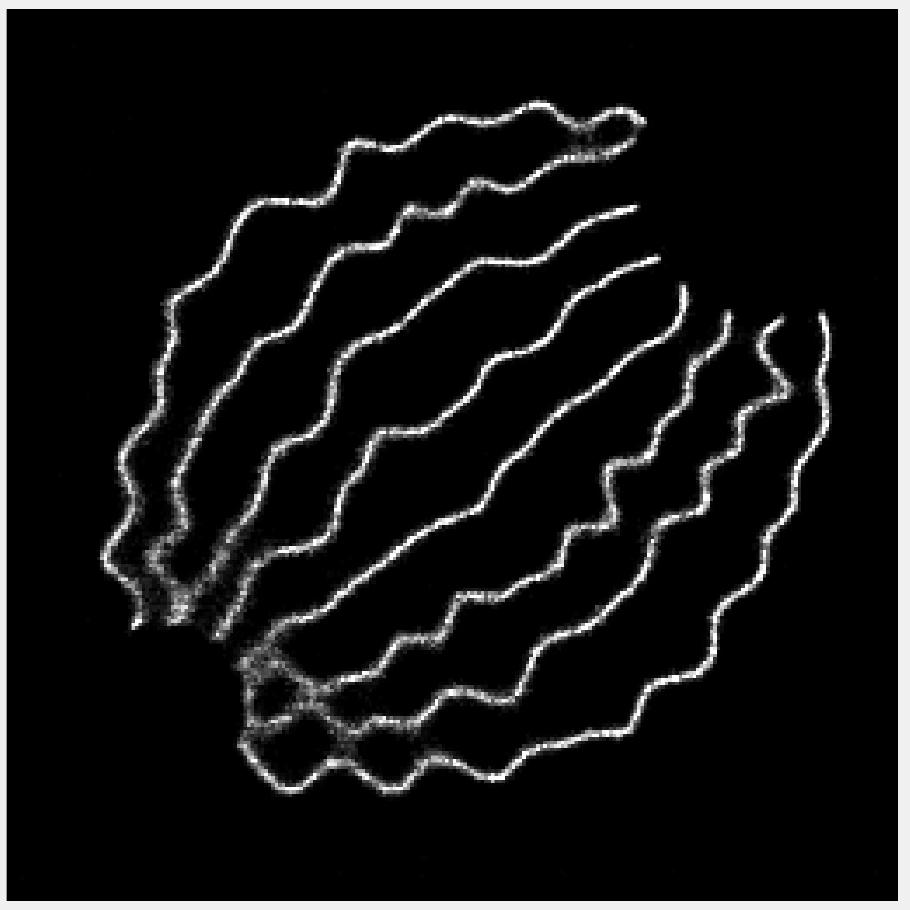}};
	\node[name=c] at (0,0) {};
	\spy[cyan] on (0.32,0.70) in node [name=d1] at (0.52,1.5);
	\spy[cyan] on (-0.22,-0.65) in node [name=d1] at (0.52,-1.5);
	\spy[cyan] on (-0.65,0) in node [name=d1] at (-0.52,-1.5);
	\draw[magenta, ultra thin] (-1,0) -- (1,0);
	\end{scope}
	\node[below of=c, align=center, node distance=2.5cm] {\textbf{DeepSTORM}};
	\end{tikzpicture}\begin{tikzpicture}
	\begin{scope}[spy using outlines={circle,black,magnification=4,size=1cm,connect spies}]
	\node[name=d] at (0,0.5) {\includegraphics[trim=0 0.5\imagewidth{} 0 0,width=2cm, clip]{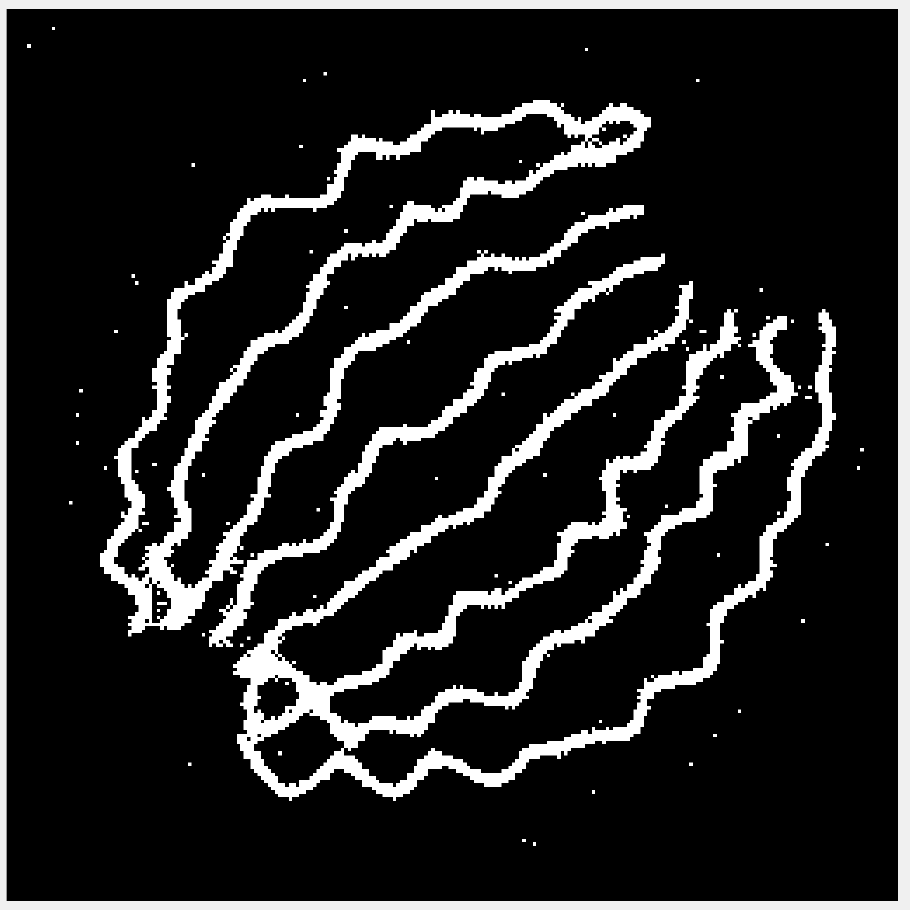}};
	\node[name=b] at (0,-0.5) {\includegraphics[trim=0 0 0 0.5\imagewidth{},width=2cm, clip]{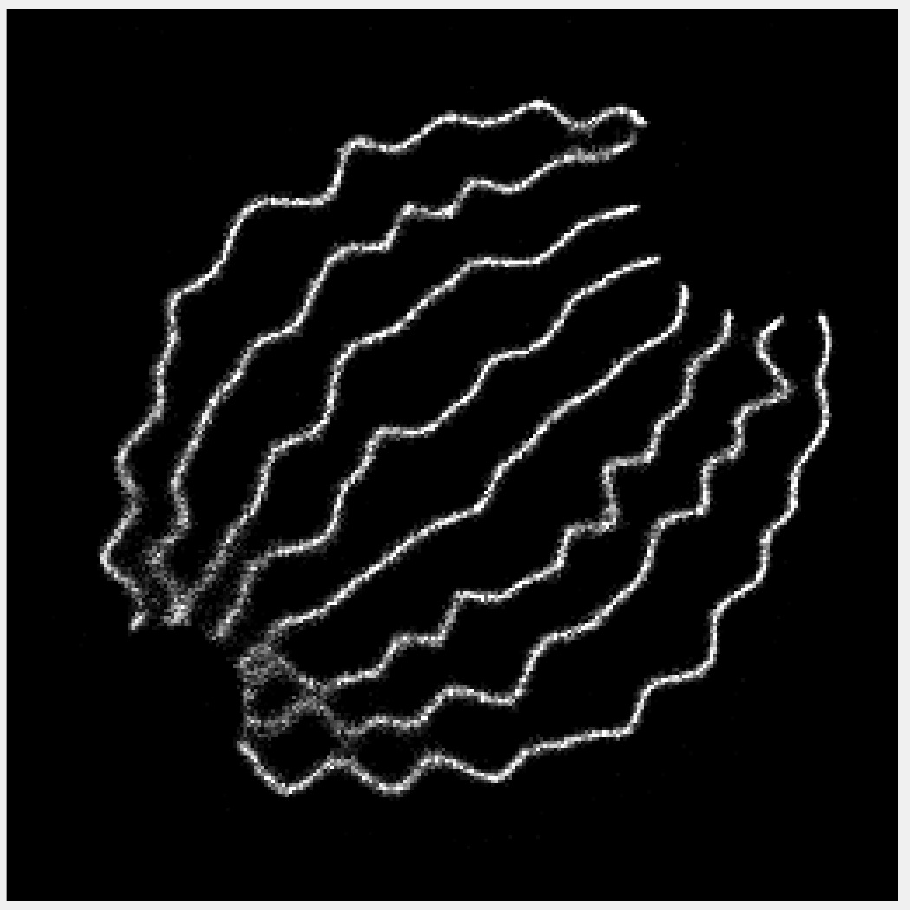}};
	\node[name=c] at (0,0) {};
	\spy[cyan] on (0.32,0.70) in node [name=d1] at (0.52,1.5);
	\spy[cyan] on (-0.22,-0.65) in node [name=d1] at (0.52,-1.5);
	\spy[cyan] on (-0.65,0) in node [name=d1] at (-0.52,-1.5);
	\draw[magenta, ultra thin] (-1,0) -- (1,0);
	\end{scope}
	\node[below of=c, align=center, node distance=2.5cm] {\textbf{DeepCEL0}};
	\end{tikzpicture}}
	\caption{\textbf{Image reconstruction results of the whole stack for the IEEE ISBI simulated dataset.}The first half of the images shows the binarized versions of the images. The second half of the images shows the normalized counterparts.}
	\label{fig:ISBI}
\end{figure}

%% file: fig4.tex
\begin{figure}[H]
    \newlength{\imagew}
    \settowidth{\imagew}{\includegraphics{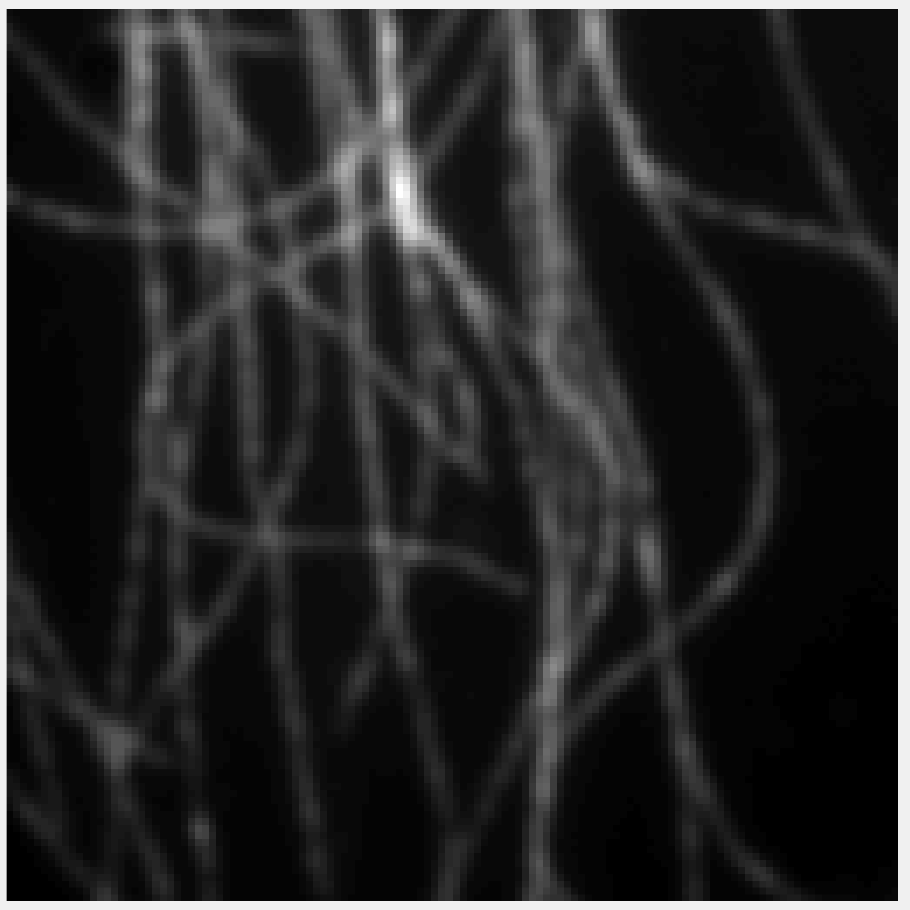}}
	\centering
	\scalebox{1.3}{\begin{tikzpicture}
	\begin{scope}[spy using outlines={circle,black,magnification=4,size=1cm,connect spies}]
	\node [name=c]{	\includegraphics[height=2cm]{img/real/LR_real.pdf}};
    \spy[yellow] on (-0.05,0.25) in node [name=c1] at (0.52,1.5);
	\spy[yellow] on (-0.64,0) in node [name=c1] at (-0.52,1.5);
	\spy[yellow] on (-0.74,-0.65) in node [name=c1] at (-0.52,-1.5);
	\end{scope}
	\node[below of=c, align=center, node distance=2.5cm] {\textbf{LR}};
	\end{tikzpicture}\begin{tikzpicture}
	\begin{scope}[spy using outlines={circle,black,magnification=4,size=1cm,connect spies}]
	\node[name=d] at (0,0.5) {\includegraphics[trim=0 0.5\imagew{} 0 0,width=2cm, clip]{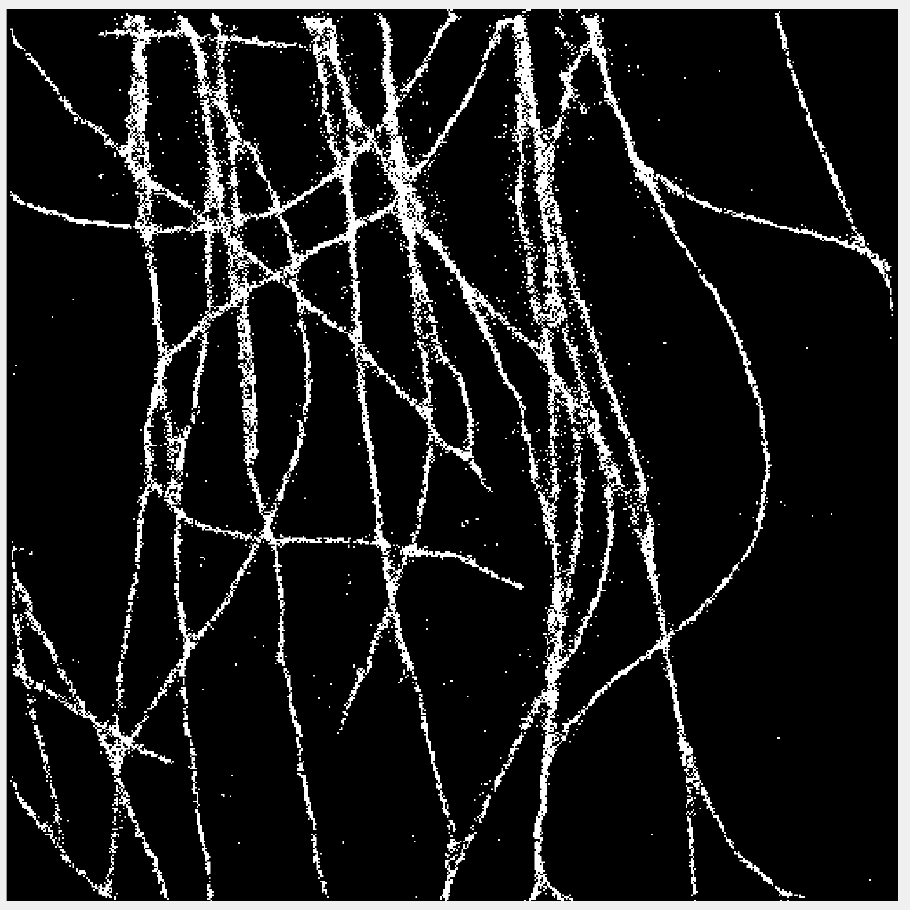}};
	\node[name=b] at (0,-0.5) {\includegraphics[trim=0 0 0 0.5\imagew{},width=2cm, clip]{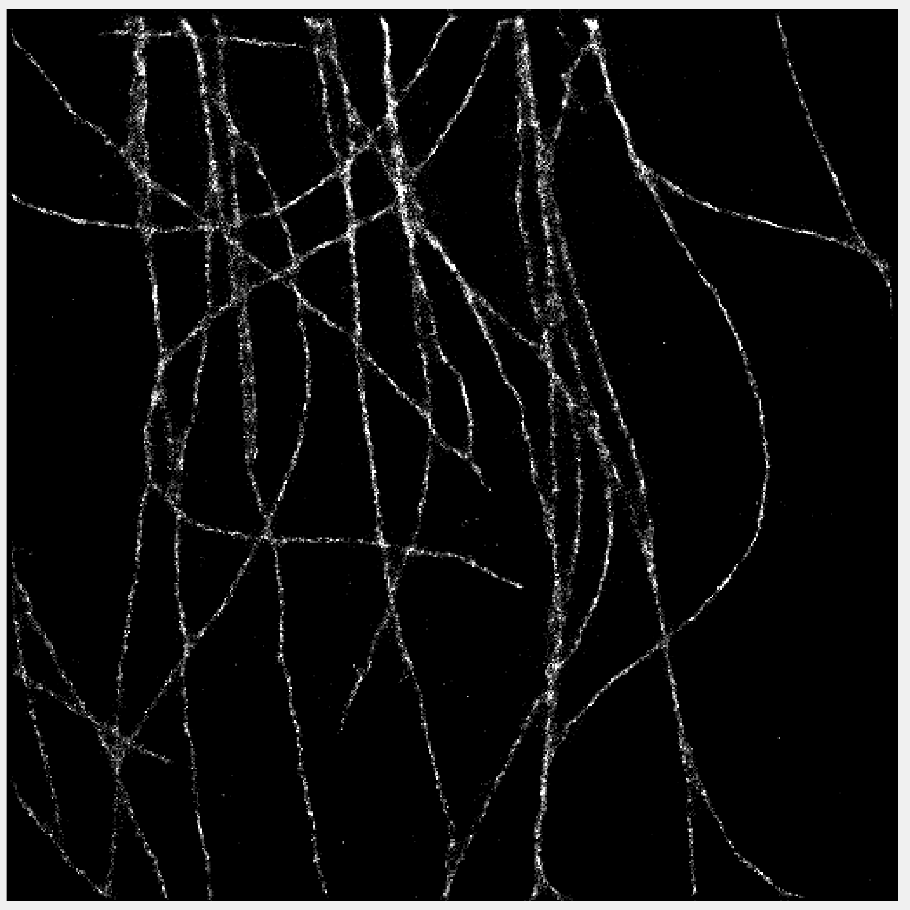}};
	\node[name=c] at (0,0) {};
    \spy[yellow] on (-0.05,0.25) in node [name=c1] at (0.52,1.5);
	\spy[yellow] on (-0.64,0) in node [name=c1] at (-0.52,1.5);
	\spy[yellow] on (-0.74,-0.65) in node [name=c1] at (-0.52,-1.5);
	\draw[green, ultra thin] (-1,0) -- (1,0);
	\end{scope}
	\node[below of=c, align=center, node distance=2.5cm] {\textbf{CEL0}};
	\end{tikzpicture}\begin{tikzpicture}
	\begin{scope}[spy using outlines={circle,black,magnification=4,size=1cm,connect spies}]
	\node[name=d] at (0,0.5) {\includegraphics[trim=0 0.5\imagew{} 0 0,width=2cm, clip]{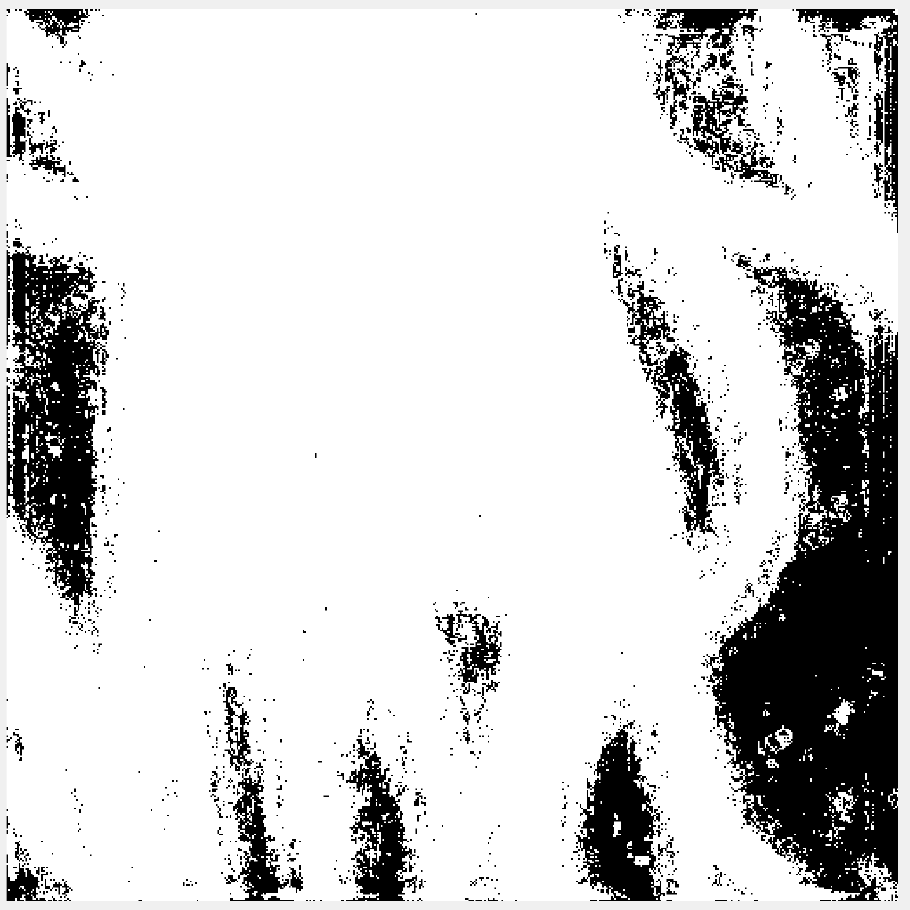}};
	\node[name=b] at (0,-0.5) {\includegraphics[trim=0 0 0 0.5\imagew{},width=2cm, clip]{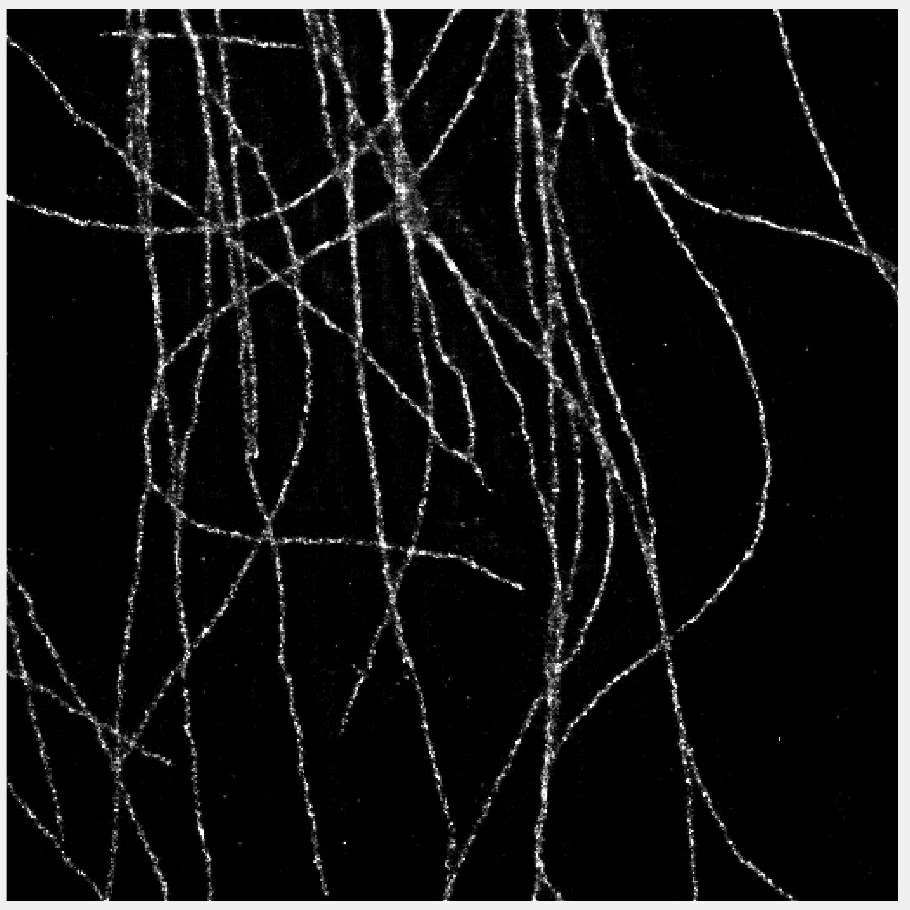}};
	\node[name=c] at (0,0) {};
    \spy[yellow] on (-0.05,0.25) in node [name=c1] at (0.52,1.5);
	\spy[yellow] on (-0.64,0) in node [name=c1] at (-0.52,1.5);
	\spy[yellow] on (-0.74,-0.65) in node [name=c1] at (-0.52,-1.5);
	\draw[green, ultra thin] (-1,0) -- (1,0);
	\end{scope}
	\node[below of=c, align=center, node distance=2.5cm] {\textbf{DeepSTORM}};
	\end{tikzpicture}\begin{tikzpicture}
	\begin{scope}[spy using outlines={circle,black,magnification=4,size=1cm,connect spies}]
	\node[name=d] at (0,0.5) {\includegraphics[trim=0 0.5\imagew{} 0 0,width=2cm, clip]{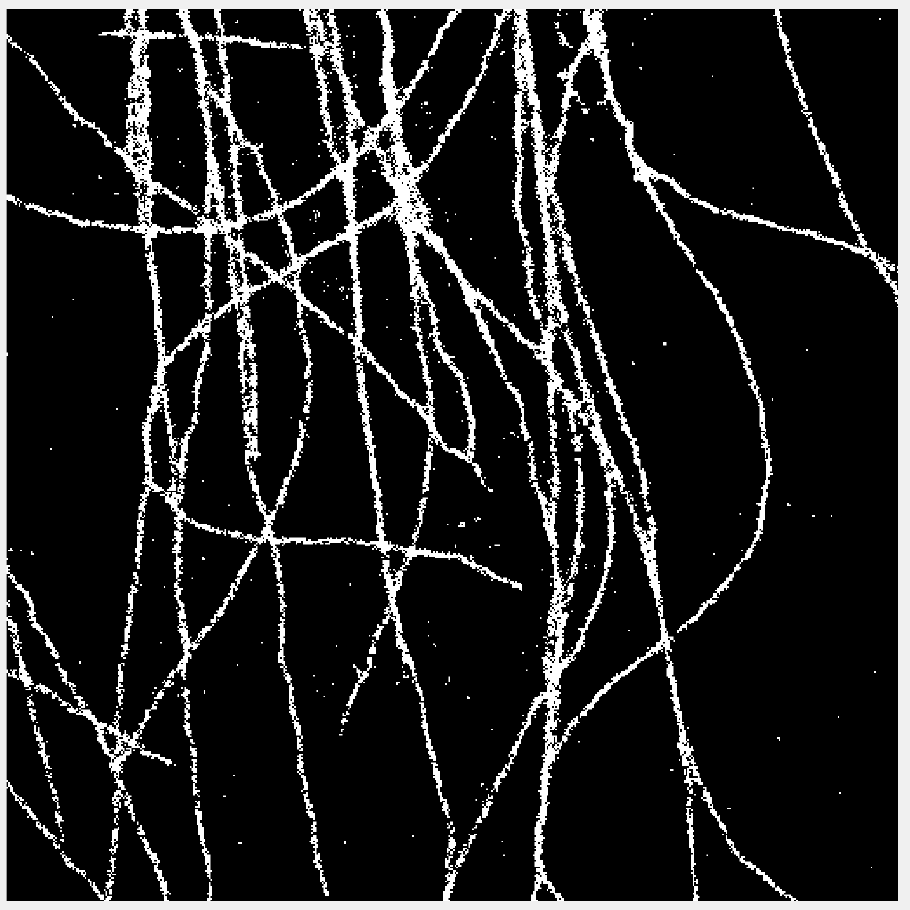}};
	\node[name=b] at (0,-0.5) {\includegraphics[trim=0 0 0 0.5\imagew{},width=2cm, clip]{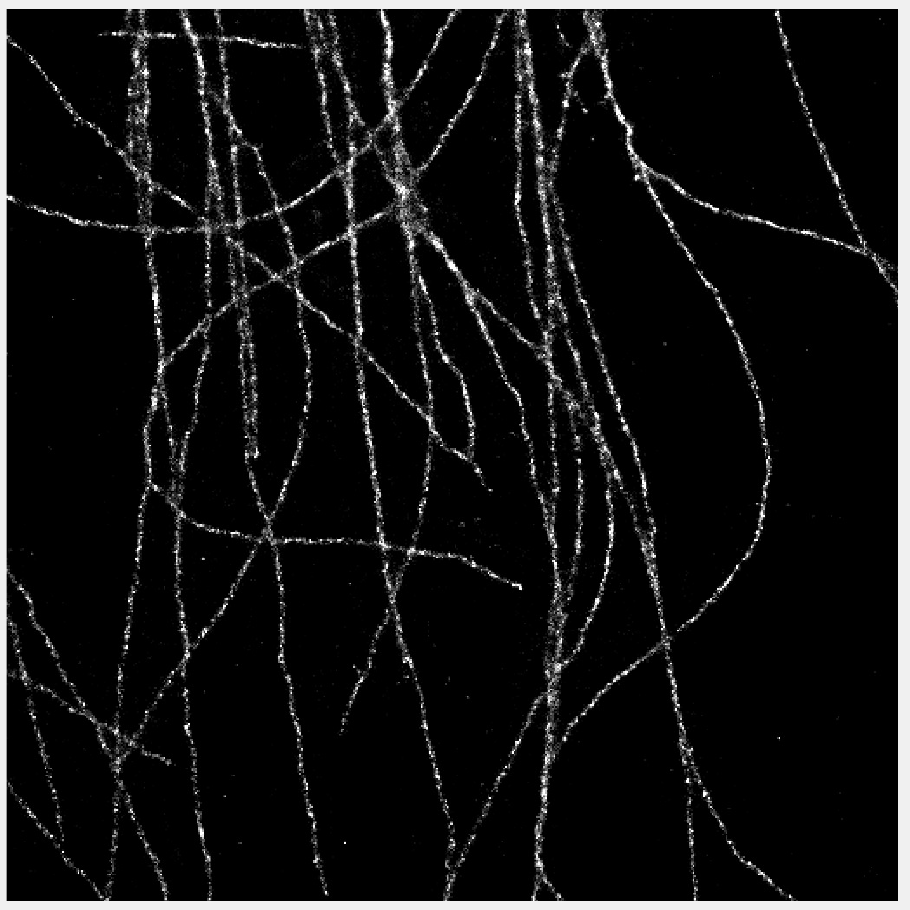}};
	\node[name=c] at (0,0) {};
    \spy[yellow] on (-0.05,0.25) in node [name=c1] at (0.52,1.5);
	\spy[yellow] on (-0.64,0) in node [name=c1] at (-0.52,1.5);
	\spy[yellow] on (-0.74,-0.65) in node [name=c1] at (-0.52,-1.5);
	\draw[green, ultra thin] (-1,0) -- (1,0);
	\end{scope}
	\node[below of=c, align=center, node distance=2.5cm] {\textbf{DeepCEL0}};
	\end{tikzpicture}}
	\caption{\textbf{Image reconstruction results of the whole stack for the IEEE ISBI tubulins dataset.} The first half of the images shows the binarized versions of the images. The second half of the images shows the normalized counterparts.}
	\label{fig:ISBI_real}
\end{figure}